%% file: 00_MAIN.tex
\documentclass[letterpaper, 10 pt, conference]{ieeeconf}  

\IEEEoverridecommandlockouts                              

\pdfoutput=1
\usepackage{textcase} 
\usepackage{graphics}
\usepackage{amsmath}
\usepackage{multirow}
\usepackage{makecell}
\usepackage[font=footnotesize,labelfont=bf]{caption}
\usepackage{subcaption}
\usepackage{rotating}
\usepackage{color}

\usepackage{afterpage}
\usepackage{float}
\usepackage{colortbl}
\usepackage{cuted}
\usepackage{xcolor}
\usepackage{graphicx}
\usepackage{subcaption}

\colorlet{lightgreen}{green!40}
\colorlet{lightyellow}{yellow!30}
\colorlet{lightred}{red!35}
\colorlet{colorSep}{blue!5}
\colorlet{lightorg}{orange!30}
\newcommand{\fst}{\cellcolor{lightred}\bf}
\newcommand{\scd}{\cellcolor{lightorg}}
\newcommand{\trd}{\cellcolor{lightyellow}}

\newcommand{\FramewkName}[1]{GaRField++#1}

\usepackage{mathtools}
\usepackage{amssymb}
\usepackage{subfig}

\title{\FramewkName{}: Reinforced Gaussian Radiance Fields for Large-Scale 3D Scene Reconstruction}

\author{Hanyue Zhang$^{1}$, Zhiliu Yang$^{1}$$^{,\;2}$$^{,\;*}$, Xinhe Zuo$^{1}$, Yuxin Tong$^{1}$, Ying Long$^{1}$, and Chen Liu$^{3}$
\thanks{$^{1}$ School of Information Science and Engineering, Yunnan University, Kunming, Yunnan 650500, China.}
\thanks{$^{2}$ Yunnan Key Laboratory of Intelligent Systems and Computing, Yunnan University, Kunming, Yunnan 650500, China.} 
\thanks{$^{3}$  Department of Electrical and Computer Engineering, Clarkson University, Potsdam, New York 13699, USA.}
\thanks{$^*$ Corresponding author, \texttt{zhiliu.yang@ynu.edu.cn}}
}

\begin{document}
\maketitle
\begin{strip}
\vspace{-6em}
\begin{center}

\centering
\setlength{\tabcolsep}{0.1em}
{\renewcommand{\arraystretch}{1.5}
\begin{tabular}{c c c c }
\includegraphics[width=0.245\linewidth]{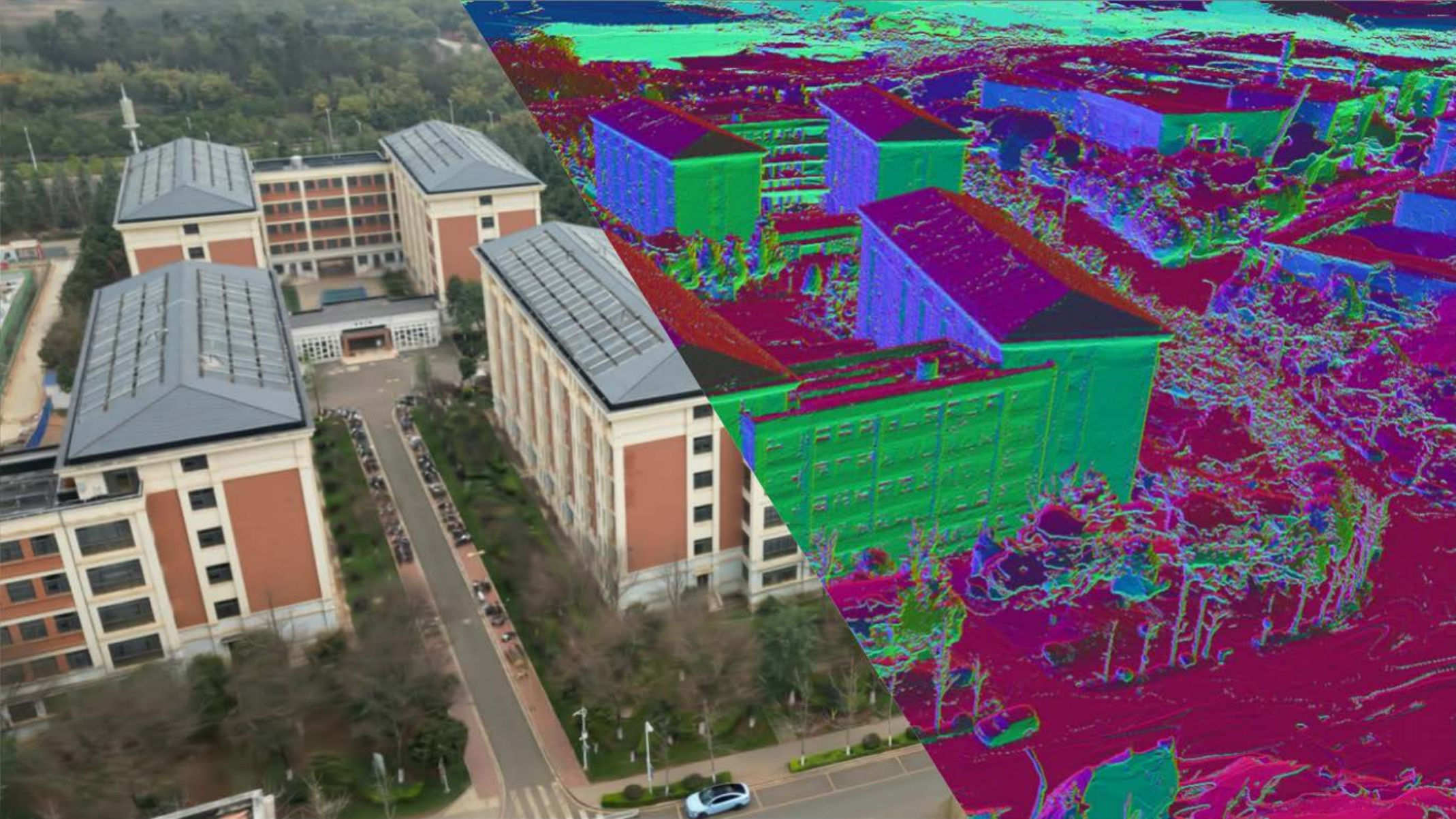} &
\includegraphics[width=0.245\linewidth]{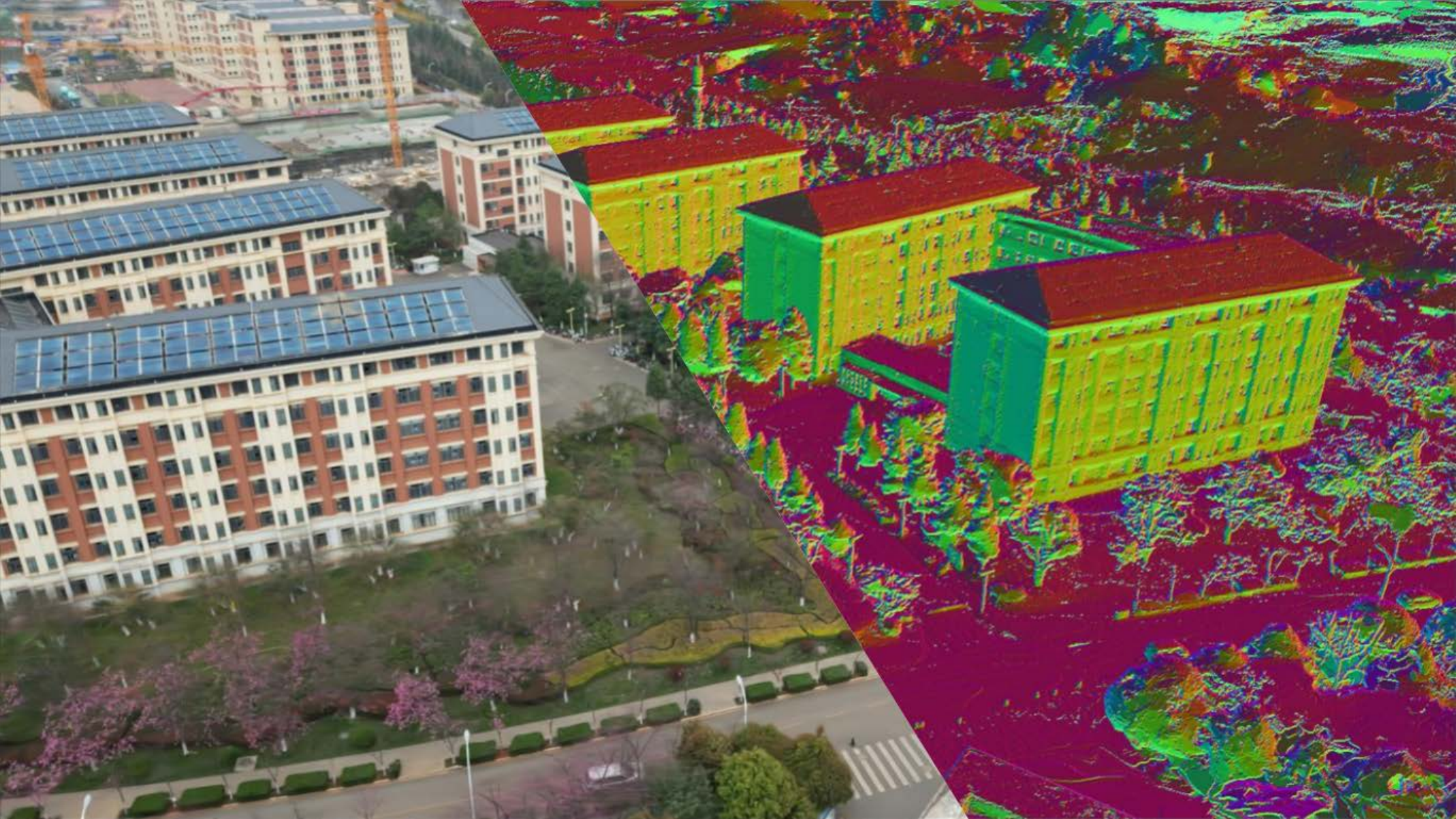} & 
\includegraphics[width=0.245\linewidth]{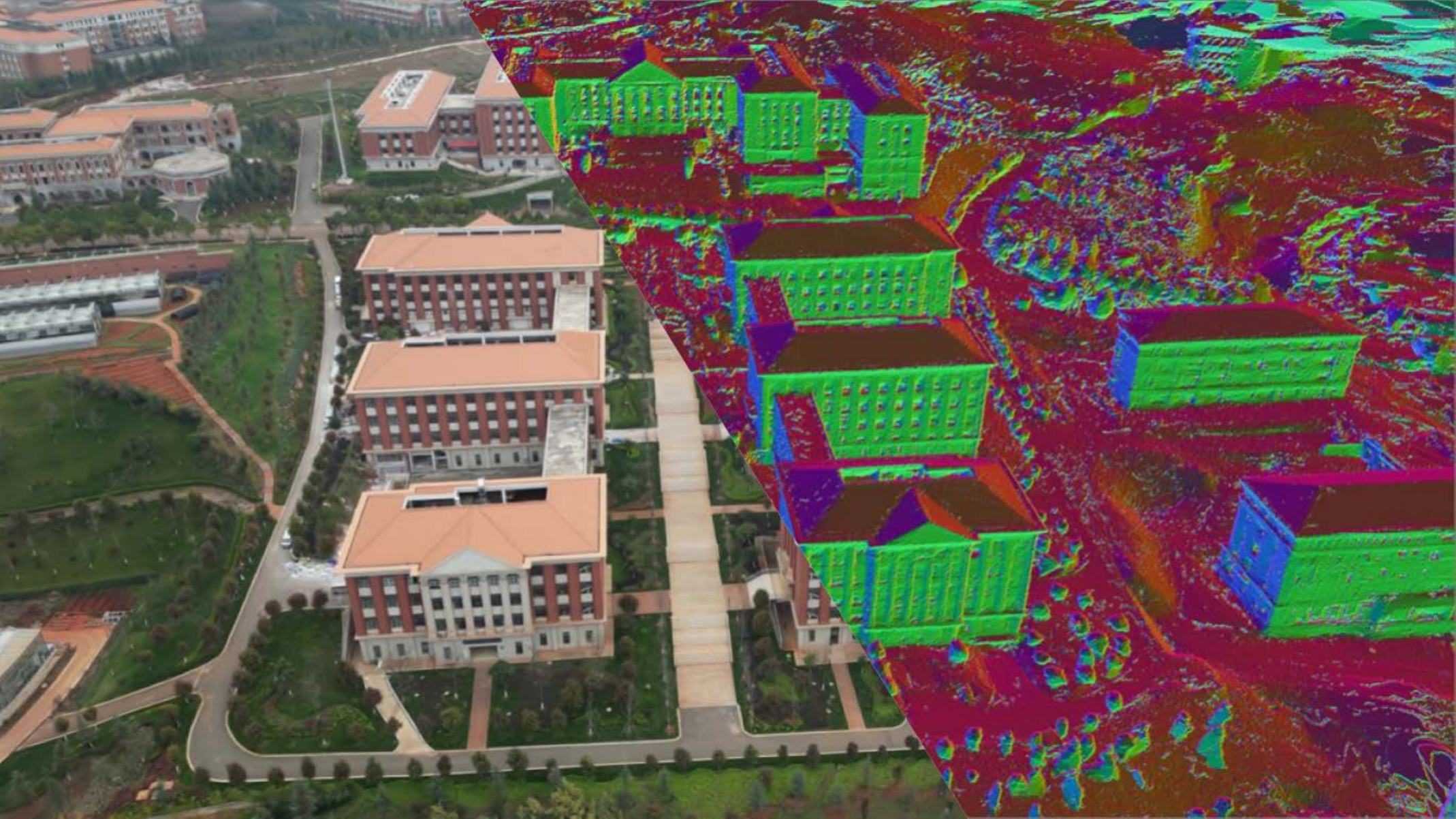}&
\includegraphics[width=0.245\linewidth]{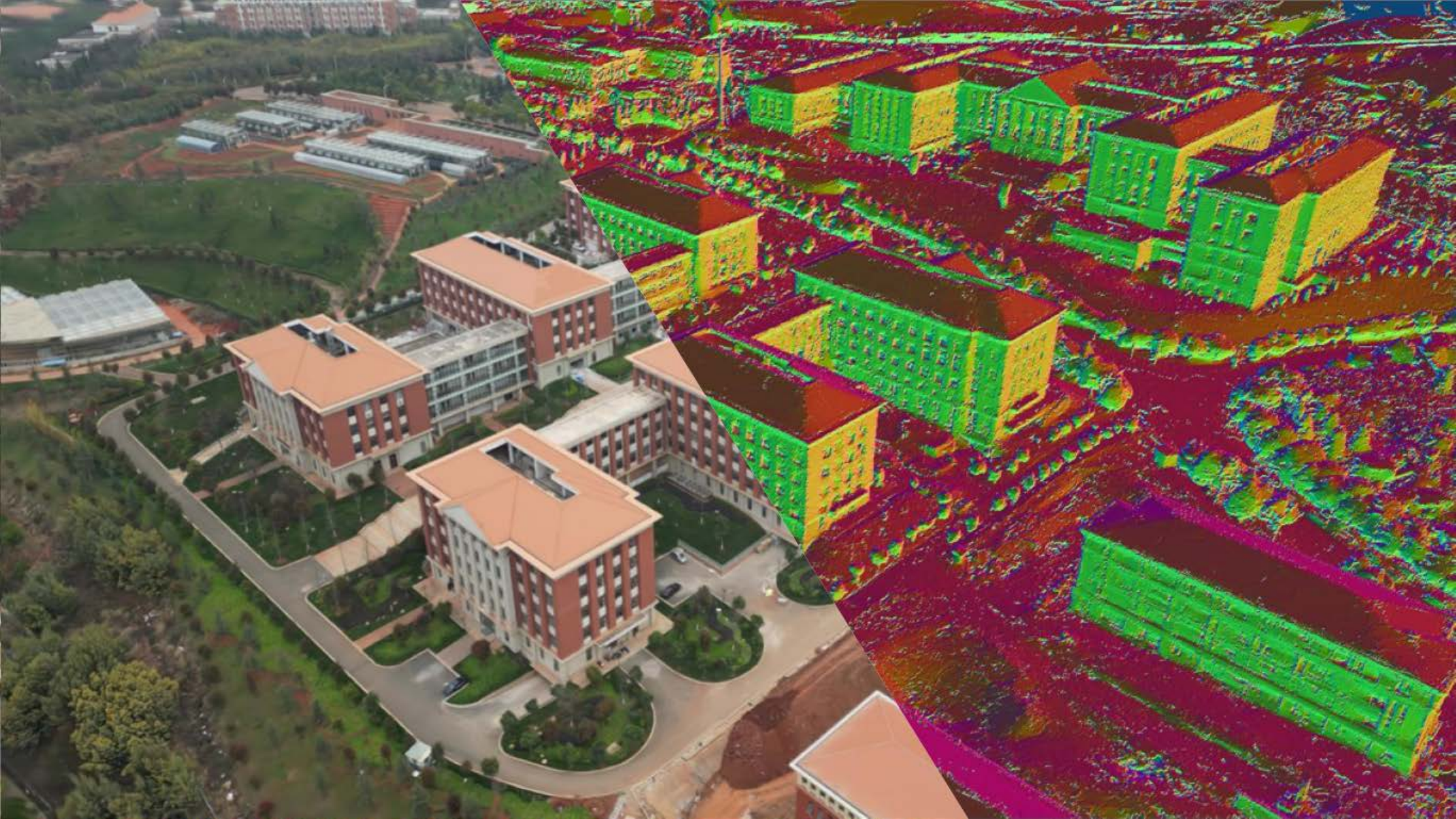}\vspace{-5pt}\\
\end{tabular}}

\footnotesize
\captionof{figure}{\textbf{Rendered RGB images and corresponding rendered depth normals from our \FramewkName{ }framework on the self-collected data.} Randomly rendered images from multiple views of the large-scale scenes are complete, smooth and detailed. This is achieved by constructing a divide-and-conquer Gaussian radiance field, which is reinforced by precisely modeling the color and opacity information and improving the training efficiency. The data is collected from the monocular camera of a DJI drone.}
\label{fig:guidance}

\vspace{-0.7cm}
\end{center}
\end{strip}
\begin{abstract}
This paper proposes a novel framework for large-scale scene reconstruction based on 3D Gaussian splatting (3DGS) and aims to address the scalability and accuracy challenges faced by existing methods. For tackling the scalability issue, we split the large scene into multiple cells, and the candidate point-cloud and camera views of each cell are correlated through a visibility-based camera selection and a progressive point-cloud extension. To reinforce the rendering quality, three highlighted improvements are made in comparison with vanilla 3DGS, which are a strategy of the ray-Gaussian intersection and the novel Gaussians density control for learning efficiency, an appearance decoupling module based on ConvKAN network to solve uneven lighting conditions in large-scale scenes, and a refined final loss with the color loss, the depth distortion loss, and the normal consistency loss. Finally, the seamless stitching procedure is executed to merge the individual Gaussian radiance field for novel view synthesis across different cells. Evaluation of Mill19, Urban3D, and MatrixCity datasets shows that our method consistently generates more high-fidelity rendering results than state-of-the-art methods of large-scale scene reconstruction. We further validate the generalizability of the proposed approach by rendering on self-collected video clips recorded by a commercial drone.
\end{abstract}

\input{01_INTRO}
\input{02_RELATE}
\input{03_METHOD}
\input{04_EXPERIM}

\input{05_CONCLUS}
\bibliographystyle{IEEEtran.bst}
\bibliography{06_BIBLIOG}

\end{document}

%% file: 01_INTRO.tex
\section{INTRODUCTION}
\label{sec:intro}
The recent advances in 3D reconstruction of large-scale urban scenes have reshaped modern society. It can serve as a visualization medium for AR/VR \cite{xu2023vr}, aerial surveying \cite{turki2022mega}, and city planning \cite{zhou2024hugs, bao20243d}, a high definition (HD) map for autonomous driving \cite{tancik2022block, yan2024gs, huang2024photo, keetha2024splatam, yugay2023gaussian, li2024sgs}, or a photorealistic simulator for unexpected cases in end-to-end autonomous driving and unmanned aerial vehicles (UAVs) \cite{zhou2024hugs, li2023read, lin2024vastgaussian, gu2024ue4}. 

The task consists of high-fidelity reconstruction and real-time rendering for large areas that typically span more than 1.5 $km^2$ \cite{turki2022mega}. In recent years, the field has been dominated by methods based on Neural Radiance Fields (NeRFs) \cite{mildenhall2021nerf}. Representative works include Block-NeRF \cite{tancik2022block}, GgNeRF \cite{xu2023grid}, Switch-NeRF \cite{zhenxing2022switch} and Mega-NeRF \cite{turki2022mega}. However, these methods still lack the fidelity in preserving details. Recently, the 3D Gaussian Splatting (3DGS) technique \cite{kerbl20233d} has gained significant attention for its outstanding performance in visual quality and rendering speed, achieving near-photorealistic rendering effects at 1080p resolution in real time. It has also been successfully applied to the reconstruction of dynamic scenes \cite{luiten2023dynamic,yang2024deformable,wu20244d} and the generation of 3D content \cite{yi2023gaussiandreamer,chen2024text}. However, the 3DGS still faces several scalability and accuracy challenges when dealing with large-scale environments.

Firstly, large-scale scenes typically encompass various objects, including the complex geometry structure such as grass, plants \cite{yu2024gaussian}, and a large area of background such as the sky and water body \cite{lin2024vastgaussian}. Traditional 3DGS-based reconstruction methods do not adequately model normal depth and opacity information. Secondly, uneven lighting conditions in large-scale scenes may lead to significant appearance differences in captured images. When dealing with these variations, 3DGS tends to generate large-size 3D Gaussians with low opacity \cite{lin2024vastgaussian}, which results in floating artifacts in novel views.
Third, optimizing the entire large-scale scene requires multiple iterations, which become extremely time consuming and unstable without the proper regularization term and loss function design \cite{huang20242d}.

Recent efforts of large-scale scene reconstruction based on the 3DGS have mitigated some of the aforementioned shortcomings. Methods like visibility-based camera selection \cite{lin2024vastgaussian}, appearance modeling \cite{yan2024street}, multimodal fusion \cite{rematas2022urban}, level of details \cite{liu2024citygaussian} etc. are proposed correspondingly to improve the rendering quality. Although these methods produce reasonable results, they are still prone to some of the blurred area in the rendered views.

We propose a reinforced Gaussian radiance field for large-scale 3D reconstruction, named \FramewkName{}. We split the large-scale scene into multiple cells by following the VastGuaussian \cite{lin2024vastgaussian}, then we implement visibility-based camera selection, relevant cameras from other cells and extended sets of the point cloud are enrolled for training to eliminate the floating artifacts. To enhance rendering fidelity, we leverage the ray-Gaussian-intersection volume rendering and improved density control strategies in the reconstruction of each cell. To mitigate uneven lighting conditions, we use a network architecture that integrates KAN \cite{liu2024kan} with convolutional neural networks (CNNs) to decouple appearance information. This color decoupling module is discarded after training to prevent impacting the rendering speed. In addition, a reinforced final loss is employed with color loss, depth distortion loss, and normal consistency loss. 


In addition to testing on the challenging public dataset, we also utilize a DJI drone (Mini 3 Pro) to capture video clips from a large-scale scene to validate the effectiveness of our approach. Our contributions are as follows.


\begin{itemize}
    \item \FramewkName{ }is the first work to leverage the ray-Gaussian intersection volume rendering and the reinforced density control strategy for the large-scale 3D reconstruction, which consistently generates more high-fidelity rendering results than state-of-the-art methods.
    \item We leverage a color decoupling module based on KAN and CNN to address the appearance variations, enhancing the fidelity of the rendering results.
    \item We exploit the depth-normal consistency to construct the regulation term for large-scale area reconstruction, to increase continuity of 3DGS optimization.
\end{itemize}

%% file: 02_RELATE.tex
\section{RELATED WORK}
\label{sec:rel}

\subsection{Rendering with Radiance Fields}
\subsubsection{Neural Radiance Fields}
Neural Radiance Fields (NeRF) \cite{mildenhall2021nerf} implicitly represents 3D scenes as a mapping from position and direction into radiance using a multi-layer perceptrons (MLPs), and achieves novel view synthesis through volumetric rendering techniques. Despite the significant progress made for 3D scene reconstruction and rendering by NeRF \cite{mildenhall2021nerf}, they still face challenges in efficiency and memory usage when dealing with large-scale scenes. To improve rendering efficiency, researchers have proposed various strategies \cite{muller2022instant, zhang2020nerf++, jaxnerf2020github}. InstantNGP \cite{muller2022instant} firstly encodes the scene into a multi-resolution hashing table. Mip-NeRF \cite{turki2022mega} enhances NeRF's representation capacity for outdoor scenes by introducing the down sampling of conical frustums. Zip-NeRF \cite{barron2023zip} employs a hexagonal sampling strategy to address aliasing issues in the rendering. 

\subsubsection{3D Gaussian Splatting}
Rendering methods based on points utilize 3D Gaussian functions as geometric primitives, achieving the rapid rendering and a scene editing ability \cite{kerbl20233d}. The 3D Gaussian Splatting (3DGS) further enhances rendering efficiency by employing optimized rasterization. Although 3DGS can produce high-fidelity 3D reconstruction results, methods such as Mip-splatting \cite{yu2024mip}, LightGaussian \cite{fan2023lightgaussian}, GSCore \cite{lee2024gscore}, Gaussianpro \cite{cheng2024gaussianpro}, Fregs \cite{zhang2024fregs}, 
 Eagles \cite{girish2023eagles}, Compact3d \cite{navaneet2023compact3d} are proposed to improve the rendering process. Motivated by the method of EWA-Splatting \cite{zwicker2001ewa}, the Mip-Splatting \cite{yu2024mip} limits the frequency of the 3D representation 
and introduces a 2D Mip filter. Eagles \cite{girish2023eagles}, Compact3d \cite{navaneet2023compact3d}, and others are committed to applying the VQ \cite{equitz1989new} trick to compress a large number of Gaussian primitives. Unlike FreGS \cite{zhang2024fregs}, C3DGS \cite{niedermayr2023compressed}, which optimizes on the software algorithms, GSCore \cite{lee2024gscore} proposes a hardware acceleration unit to optimize the 3DGS pipeline in the rendering of the radiance field. GaussianPro \cite{cheng2024gaussianpro} introduces an innovative paradigm for joint 2D-3D training to reduce the dependence on SfM initialization.

\begin{figure*}[ht]
    \centering
    \includegraphics[width=0.99\linewidth]{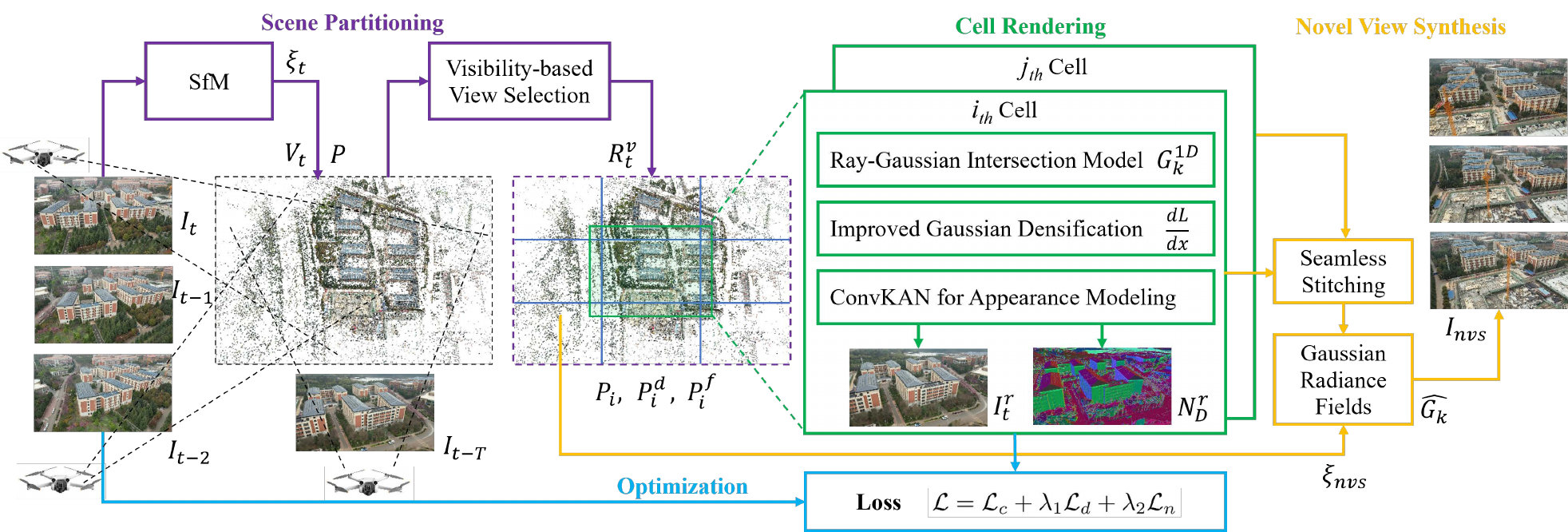}
    \caption{\textbf{Overview of our \FramewkName{ } framework.} \textbf{Scene Partitioning:} We implement a sparse reconstruction based on the Structure-from-Motion (SfM) method, generating a point cloud and estimating the initial camera pose for each image. Concurrently, we performed Manhattan alignment on the point cloud. Subsequently, we employ a coordinate-based regionalization and a visibility-based view selection strategy to split the point cloud. \textbf{Cell Rendering: } By leveraging the ray-Gaussian intersection model, enhanced Gaussian density control, and convolution KAN (Kernelized Attention Network)-based decoupled appearance modeling, we obtained the reconstruction results for each partition. \textbf{Optimization: } We employ a newly constructed loss function to optimize the training process. This loss function encompasses depth distortion loss, normal consistency loss, and color loss, thereby enhancing the accuracy and efficiency of large-scale reconstruction.
    \textbf{Novel View Synthesis: } we seamlessly stitched together the separate Gaussian fields from various cells to obtain a complete Gaussian field for the large-scale scene. This step enables the entire large-scale area model to support cross-border rendering, providing the possibility for the generation of novel view synthesis.}
    \label{fig:overview}
\end{figure*}

\subsection{Large-scale Scene Reconstruction}
The neural rendering and the 3DGS-based rendering are naturally extended to the domain of large-scale scene reconstruction. Block-NeRF \cite{tancik2022block} divides large scenes into blocks and introduces appearance embeddings, learned pose refinement, and controllable exposure for the training of each individual block. Mega-NeRF \cite{turki2022mega} analyzes the data visibility of large-scale scenes, thereby proposes a sparse network structure where parameters are dedicated to different areas of the scene. Urban Radiance Fields \cite{rematas2022urban} utilizes LiDAR and 2D optical flow data for large-scale scene reconstruction. Switch-NeRF \cite{mi2023switchnerf} introduces a Mixture of Experts (MoE) system for end-to-end large-space modeling. A 3D point is assigned to an expert through a gating network, and the final rendering outcome is determined by the combined output of the expert and the gate value. VastGaussian \cite{lin2024vastgaussian} and CityGaussian \cite{liu2024citygaussian} are representative works that take advantage of 3DGS for scalability and rendering fidelity of large-scale scene reconstruction. 
Additionally, DrivingGaussian \cite{zhou2024drivinggaussian} and StreetGaussians \cite{yan2024street} aim at reconstructing large-scale dynamic scenes in autonomous driving using multi-modal data. StreetGaussians \cite{yan2024street} uses Fourier transforms to effectively represent the temporal changes of spherical harmonics. DrivingGaussian \cite{zhou2024drivinggaussian} leverages the LiDAR priors and employs multi-frame multi-view data for hierarchical scene modeling. 3DGS-Calib \cite{herau20243dgs} introduces LiDAR point clouds as reference points for Gaussian positions to construct a continuous scene representation. 

While the aforementioned studies have effectively improved the rendering quality in the large-scale scene reconstruction compared to the methods proposed before inventing the NeRFs and the 3DGS, there is the space for improving the rendering precision of geometric structure and large homogeneous areas.

%% file: 03_METHOD.tex
\section{METHODOLOGY}
\label{sec:method}
Our \FramewkName{ }framework processes the input images through a structure-from-motions module, a scenes partitioning, a cells rendering, and a seamless stitching to construct a reinforced Gaussian radiance field, which gives its capability to synthesize photorealistic views. The overview of the entire framework is shown in Fig. \ref{fig:overview}.




\subsection{Scenes Partitioning}
We employ a divide-and-conquer strategy similar to \cite{lin2024vastgaussian} and \cite{liu2024citygaussian}, divide the large-scale scene into multiple cells, then render each cell independently.

\subsubsection{\textbf{Sparse Reconstruction}}
The input images of the scene are denoted as $\{I_t|t = 1, 2, ...,T\}$. Then the Structure-from-Motion (SfM) method, COLMAP \cite{schonberger2016structure}, is adopted to generate a sparse point cloud $P$, and the initial camera pose $\xi_{t}$ is estimated for each image $I_t$. The camera views are defined as $V_t = \{I_t, \xi_{t}\}$. The Z axis of the point cloud $P$ is adjusted to be perpendicular to the ground plane by performing Manhattan world alignment \cite{lin2024vastgaussian}. 

\subsubsection{\textbf{Visibility-based View Selection}}
The best illumination condition and geometry visibility can be obtained by applying the coverage-wise point selection strategy, and details of the view selection are given below.
\begin{itemize}
    \item \textbf{Coordinate-based Regionalization:} The large-scale scene is first divided into $N$ cells and we distribute parts of the point cloud to a specific cell. The point cloud within a cell is defined as $\{P_i | i = 1, 2, 3, ..., N\}$.
    \item \textbf{Point Clouds Extension:} Boundaries of the cell $i$ are expanded to enroll the common views between adjacent cells. The original bounded area of cell $i$ is $L_i^W \times L_i^H$, which now extends to $i$ is $(1 + \beta) L_i^W \times (1+\beta) L_i^H$ by a certain percentage $\beta$. The set of point clouds $P_i$ is slightly dilated to $P_i^d$.
    \item \textbf{Cameras and Points Selection for Data Partitioning: }
    Given a cell $i$, the camera views from the adjacent cell $j$ is enrolled by checking the visibility criterion $R_t^v$, which is calculated by the following equation:
    \begin{equation}
    \label{equ:ratio_vis}
        R_t^v = \{\frac{A_{proj}}{A_t^j}|A_t^j = W_t^j \times H_t^j\}
    \end{equation}
    Where $A_{proj}$ is the projected area of $i_{th}$ cell in image $I_t^j$ and $A_t$ is the area of pixels in image $I_t^j$ by multiplying the width of the image $W_t^j$ and height $H_t^j$. Cameras whose $R_t^v$ is larger than a predefined threshold \cite{lin2024vastgaussian} are selected to join the cell $i$. And more point cloud from the adjacent cell $j$ is selected in the current partition, only if those points can be observed from the newly added camera views $V_t$. The final point cloud inside a cell $i$ is further extended to $P_i^{f}$.
\end{itemize}

\subsection{Cells Rendering}
The previous step produces the best point set, $P_i^f$, for modeling one of the partitions of large-scale areas, which represents a coarse description of the geometry distribution. Here, we further correlate these points with Gaussian primitives \cite{kerbl20233d}. And our \FramewkName{ }framework strengthens the radiance fields made up of Gaussian primitives with the following three reinforcements.
\subsubsection{\textbf{Ray-Gaussian Intersection Model \& Improved Gaussian Density Control}}
The sparse point clouds of the scene is further depicted with a set of 3D Gaussian primitives $\{G_k | k = 1, \dots, K\}$ correspondingly. The properties of each 3D Gaussian $G_k$ are parameterized by view-dependent color $ \mathbf{c}_k \in \mathbb{R}^{3 \times 1}$, opacity $\alpha_k \in [0, 1]$, center $\mathbf{u}_k \in \mathbb{R}^{3 \times 1}$, scale $\mathbf{s} \in \mathbb{R}^{3 \times 1}$, and rotation $\mathbf{R} \in \mathbb{R}^{3 \times 3}$.

The Gaussian primitive $G_k$ of any point $ \mathbf{x} \in \mathbb{R}^{3 \times 1}$ is depicted as:
\begin{gather}    
\label{equ:gs_prim}
    G_{k}(\mathbf{x}) = \alpha_ke^{-\frac{1}{2}(\mathbf{x}-u_{k})^{T}\Sigma_{k}^{-1}(\mathbf{x}-u_{k})}
\end{gather} 
Different from original 3DGS \cite{kerbl20233d} method which projects Gaussian balls into 2D screen space and examine the Gaussian in 2D, ray-Gaussian intersection \cite{yu2024gaussian} is utilized here to convert 3D Gaussians at any point $\mathbf{x}$ into a 1D Gaussian $G_{k}^{1D}(\mathbf{x})$. For a given camera pose $\xi_t$ of the image $I_t$, the contribution of Gaussian along its ray is defined as $\psi(G_k^{1D},\xi_t)$. Then the color of a pixel $p_v$ in $I_t$ is rendered via alpha blending along the camera ray:
\begin{gather}    
\label{equ:alpha_bled}
\textbf{c}(p_v) = \sum_{k=1}^{K} \textbf{c}_k \alpha_k \psi\left(\mathcal{G}_k^{1D}, \xi_t\right) \prod_{j=1}^{k-1} \left(1 - \alpha_j \psi\left(\mathcal{G}_j^{1D}, \xi_t\right)\right)
\end{gather} 
By utilizing the ray tracing volume rendering in Equation (\ref{equ:alpha_bled}), the opacity along the ray is monotonically increasing until it reaches the maximal value.

Motivated by \cite{yu2024gaussian}, an improved Gaussian densification strategy is used, in addition to the classical cloning or splitting, to handle
areas that are overly blurred. To enlarge the gradients values, the magnitude of view position gradient is redesigned as:
\begin{gather}    
\label{equ:GS_densify}
\frac{d L}{d\textbf{x}} = \sum_{v} \left\| \frac{d L}{d p_{v}} \frac{d p_{v}}{d \textbf{x}} \right\|
\end{gather} 
where $\textbf{x}$ is the center of Gaussian, $p_v$ is the pixels, and $\frac{dL}{d\textbf{x}}$ is the position gradient of 3DGS \cite{kerbl20233d}. Accumulating the norms $\|\cdot\|$ prevents the gradient signals from different pixels to negate each other. The densification strategy in our framework is executed at every certain iterations during the rendering.

\subsubsection{\textbf{ConvKAN-based Decoupled Appearance Modeling}}
To address the potential inconsistency between geometry and lighting in the rendering process, decoupled appearance modeling is required. The VastGaussian \cite{lin2024vastgaussian} utilizes a small CNN to predict the colors and illuminations of the images. Inspired by the Kernelized Attention Network (KAN) \cite{liu2024kan, bodner2024convolutional}, our decoupling network is designed by inserting KAN into CNNs. Replacing part of traditional convolution operations with KAN can improve rendering quality while keeping the model parameters almost unchanged.
\begin{figure}[ht]
    \centering
    \includegraphics[width=0.99\linewidth]{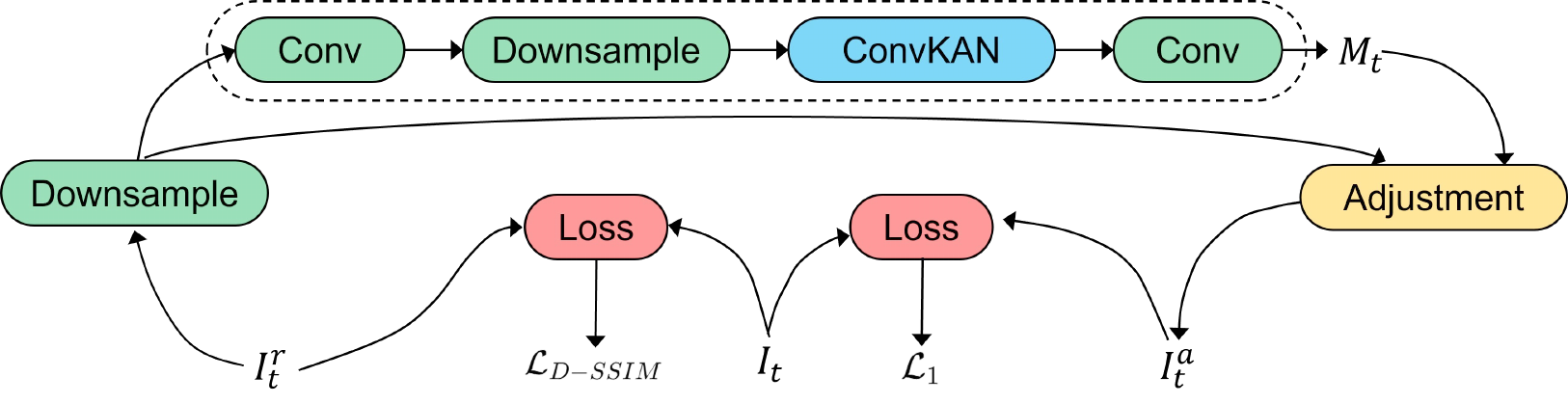}
    \caption{\textbf{Architecture of our ConvKAN-based decoupled appearance modeling.} 
}
    \label{fig:KAN}
\end{figure}

As shown in Fig. \ref{fig:KAN}, our decoupled appearance model consists of an initial convolutional layer that processes the initial input to extract preliminary features, a downsampling block, and a final convolutional layer where KAN replaces traditional convolution operations. The role of the downsampling block is to progressively downsample the feature maps, reducing the spatial resolution. The convKAN layer further processes the downsampled features and finally produces output through a sigmoid activation layer, with values ranging between 0 and 1. This color decoupling module is discarded after training, and thus it will not impact the rendering time.

\subsubsection{\textbf{Optimization}}
Rudimentary photo-metric loss is not reliable and effective for modeling large-scale reconstruction. Motivated by the regularization terms in 2DGS \cite{huang20242d} and GOF \cite{yu2024gaussian}, we optimize Gaussian model of $i_{th}$ cell with the following loss function:
\begin{gather}    
\label{equ:final_loss}
\mathcal{L} = \mathcal{L}_{c} + \lambda_1 \mathcal{L}_{d} + \lambda_2 \mathcal{L}_{n}
\end{gather}    
$\mathcal{L}_{d}$ is the depth distortion loss proposed by 2DGS \cite{huang20242d}. $\mathcal{L}_{n}$ is normal consistency loss, the normal $\textbf{N}_D$ is estimated by the gradient of the depth map $D_t$. $\mathcal{L}_{c}$ is a RGB loss from 3DGS \cite{kerbl20233d}, which is defined as follow:
\begin{gather}    
\label{equ:LC_loss}
\mathcal{L}_{c} = \mathcal{L}_{1}\left(I_t^{a},I_t\right) + \lambda_3 \mathcal{L}_{D-SSIM} \left(I_t^r, I_t\right)
\end{gather}
As shown in Fig. \ref{fig:KAN}, the \(\mathcal{L}_{D-SSIM} \) metric predominantly penalizes deviations in structural integrity, and its application to the comparison between the rendered image \(I_t^r \) and the original image \( I_t \) ensures a high degree of appearance alignment between \(I_t^a \) and \( I_t\). Meanwhile, the task of recognizing appearance features is fulfilled by embeddings \( L_t \) and our ConvKAN-based network. Furthermore, the loss function \( \mathcal{L}_{1} \) is utilized to address the appearance discrepancies between the rendered image \( I_t^a \) and the actual scene image \( I_t \), accommodating ground truth images that may exhibit subtle variations in appearance compared to other images. After training, the rendered image \(I_t^r\) is expected to maintain a consistent appearance with other images, enabling the Gaussian radiance field to learn the average appearance characteristics across all input views, as well as the precise geometric structure.
\subsection{Seamless Stitching \& Novel view Synthesis}
The Gaussian radiance fields within each cell is well-trained, and the Gaussian points outside the original boundary presented by $P_i$ (before the boundaries extension step) is cut out for seamless merging. Then we directly stitch different cells together, and the entire large-scale area model can now support cross-boarder rendering for novel view synthesis. Given a random camera pose $\xi_{nvs}$, the novel view $I_{nvs}$ can be rendered. To be noticed, $I_{nvs}$ are neither seen in the training datasets nor the testing dataset. 

%% file: 04_EXPERIM.tex
\section{Experiments}\label{sec:exp}
\subsection{Experimental Setup}
\subsubsection{\textbf{Dataset and Metrics}}
The experiments are conducted across five large-scale scenarios: the Rubble and the Building from the Mill-19 dataset \cite{turki2022mega}, the Residence from the UrbanScene3D dataset \cite{UrbanScene3D}, the small\_city, which is a synthetic scene from the MatrixCity dataset \cite{li2023matrixcity}, and Campus-YNU dataset collected by ourselves. The Campus-YNU dataset covers a region around $1 \text{km} \times 1 \text{km}$, which is captured simply using a DJI drone (Mini 3 Pro). SSIM, PSNR, and LPIPS \cite{zhang2018unreasonable} architecture serve as our evaluation metrics to quantitatively analyze the rendering results. All experiments were conducted on NVIDIA L40 GPUs with 48 GB memory for each card.

\subsubsection{\textbf{Implementations and Baselines}}

\begin{figure*}[ht]
  \centering
  \setlength{\tabcolsep}{0.7pt}
     \vspace*{5pt}

  \begin{tabular}{ccccc}
    \multicolumn{1}{c}{Ground Truth} & \multicolumn{1}{c}{GP-NeRF \cite{xu2023grid}} & \multicolumn{1}{c}{3DGS \cite{kerbl20233d}} & \multicolumn{1}{c}{CityGS \cite{liu2024citygaussian}} & \multicolumn{1}{c}{Ours} \vspace{-0.1em} \\
     \makecell{\includegraphics[width=.196\linewidth]{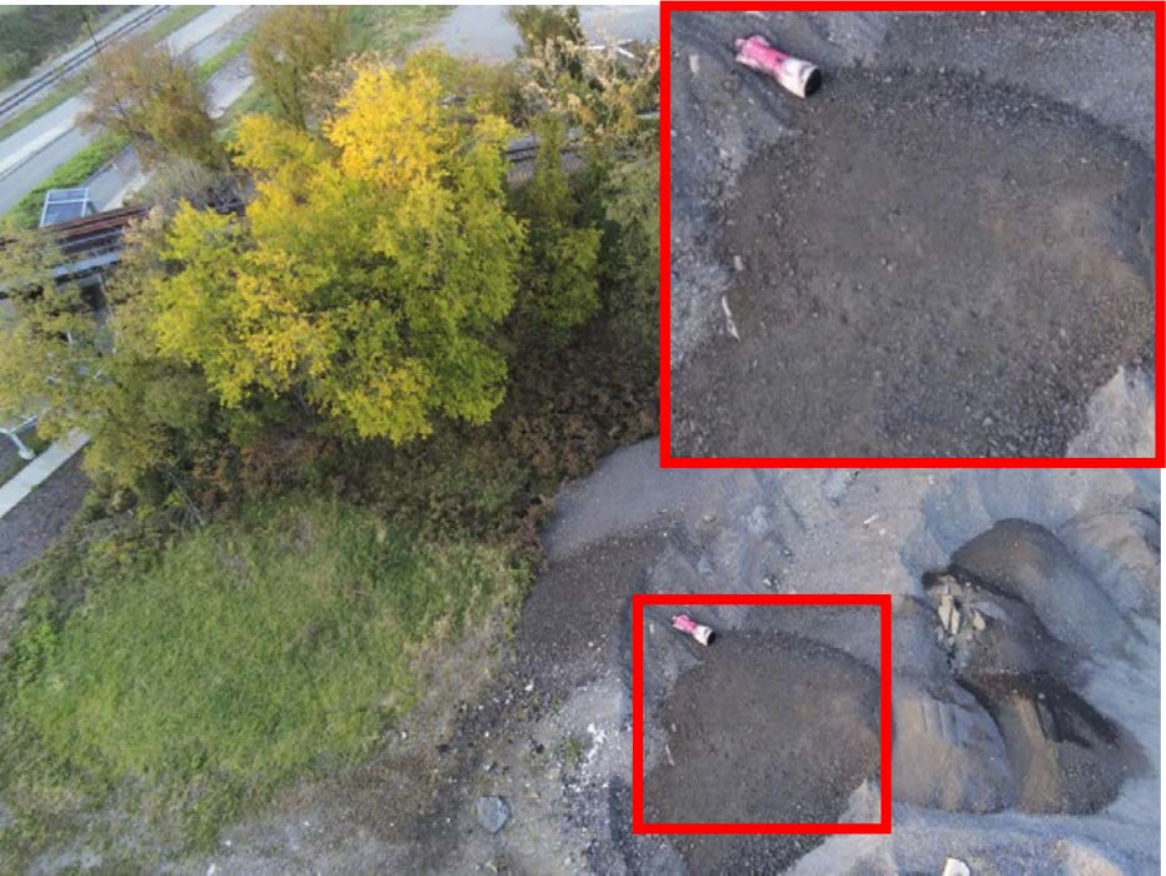}} & 
    \makecell{\includegraphics[width=.196\linewidth]{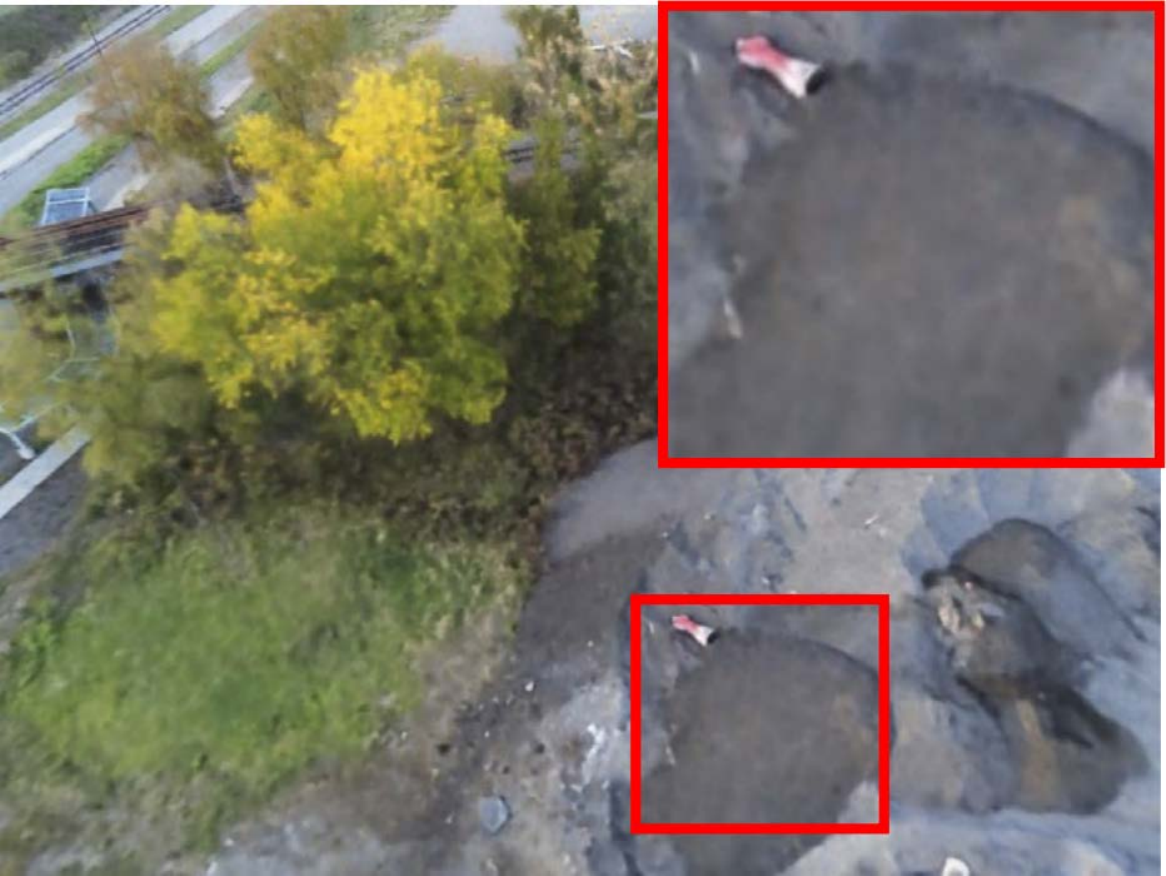}} &
    \makecell{\includegraphics[width=.196\linewidth]{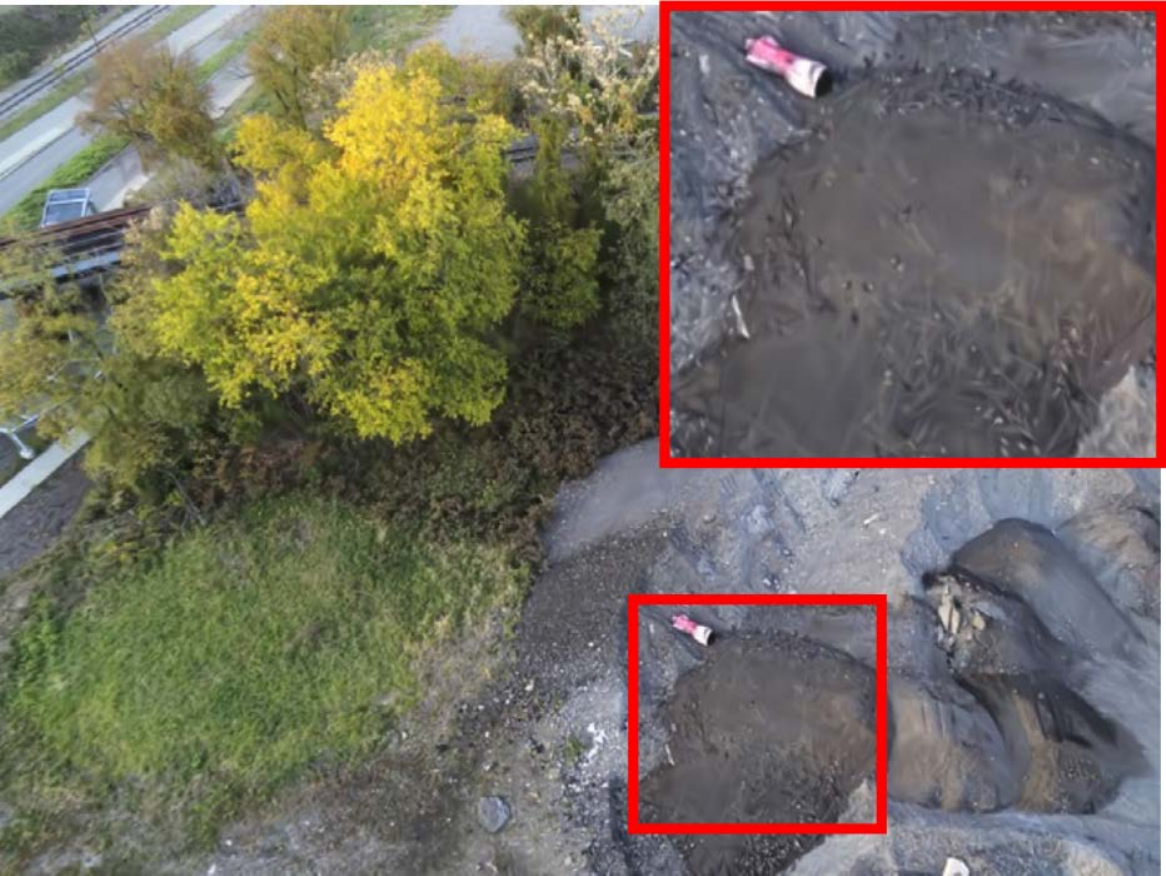}} & 
    \makecell{\includegraphics[width=.196\linewidth]{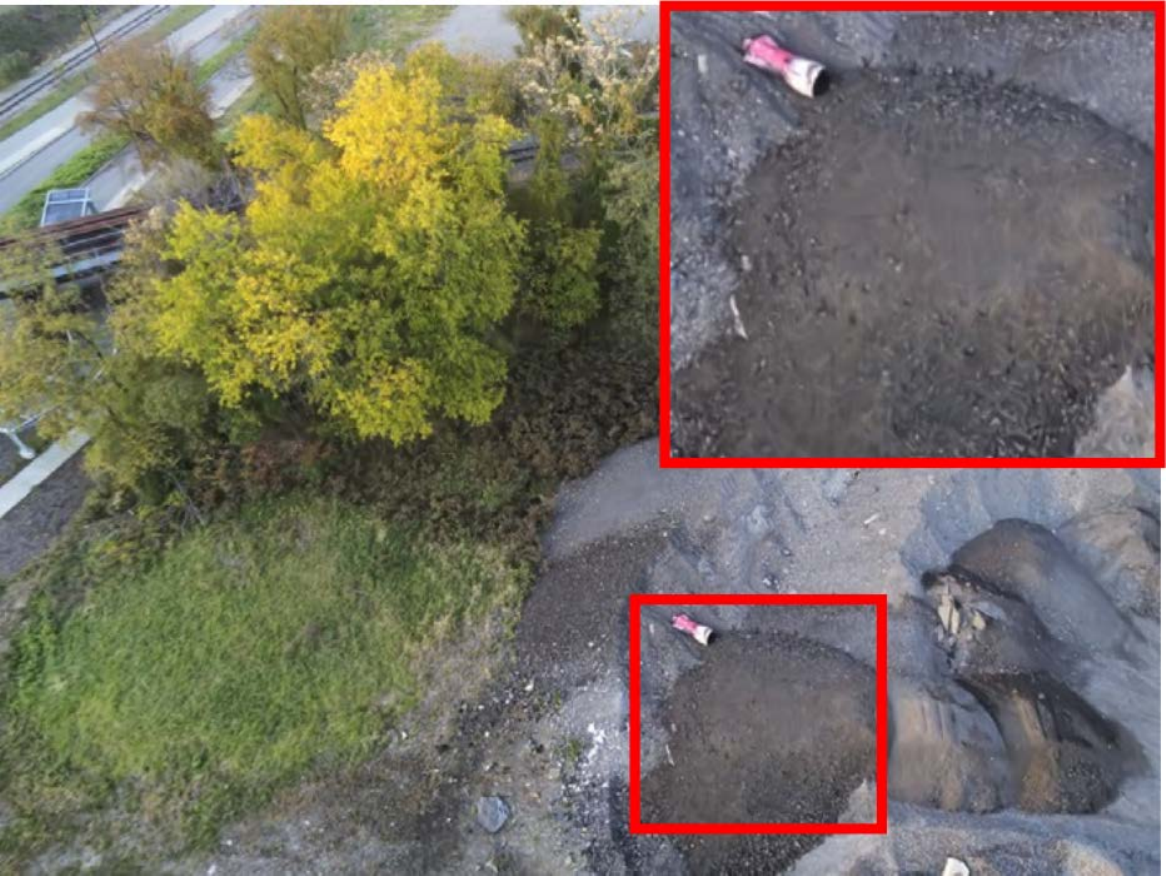}} & 
    \makecell{\includegraphics[width=.196\linewidth]{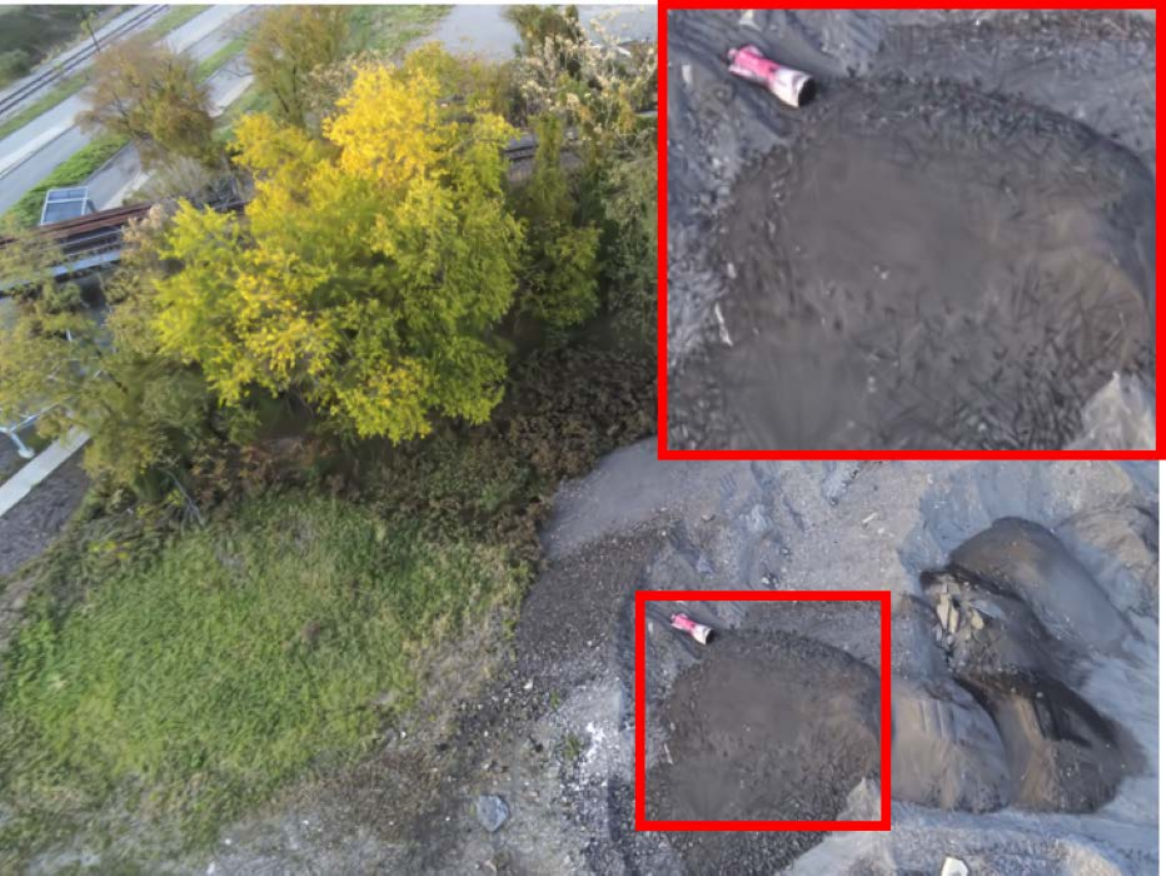}}\vspace{-2.5pt} \\
    \makecell{\includegraphics[width=.196\linewidth]{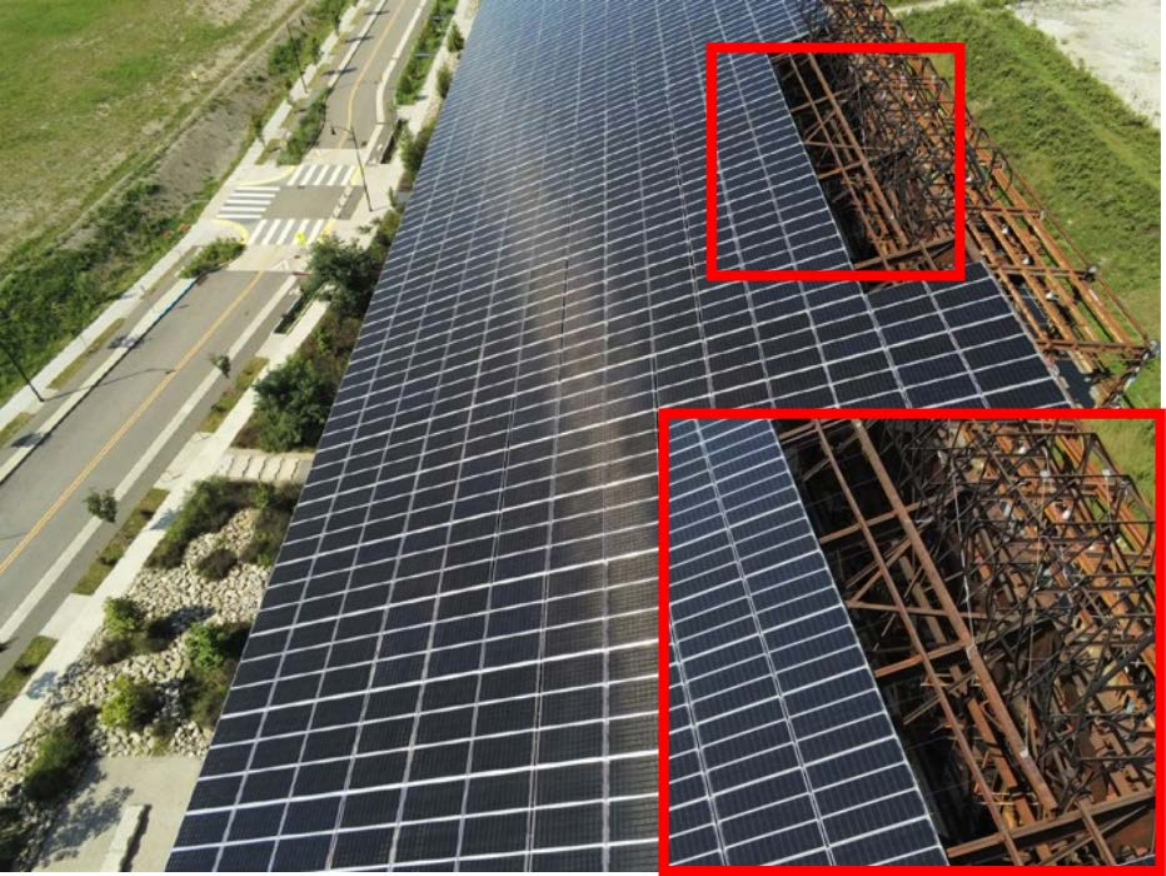}} & 
    \makecell{\includegraphics[width=.196\linewidth]{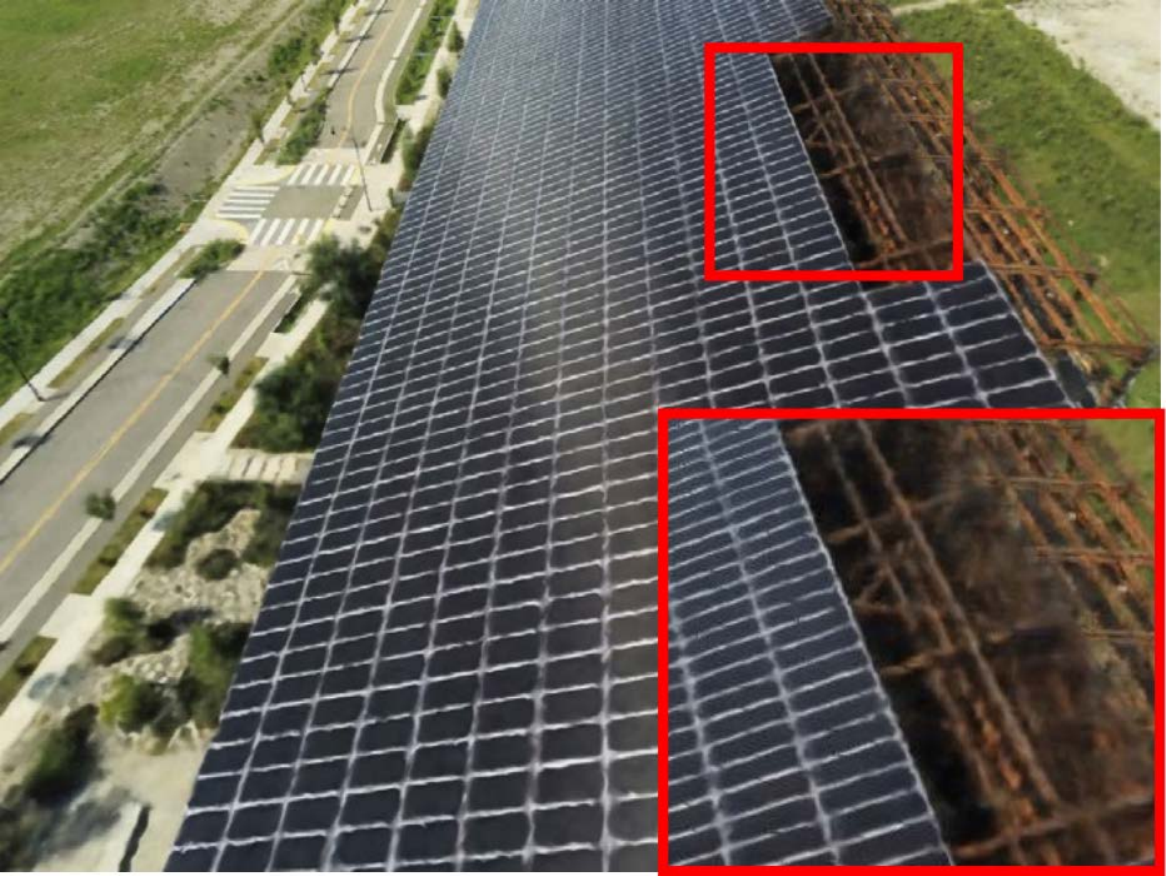}} &
    \makecell{\includegraphics[width=.196\linewidth]{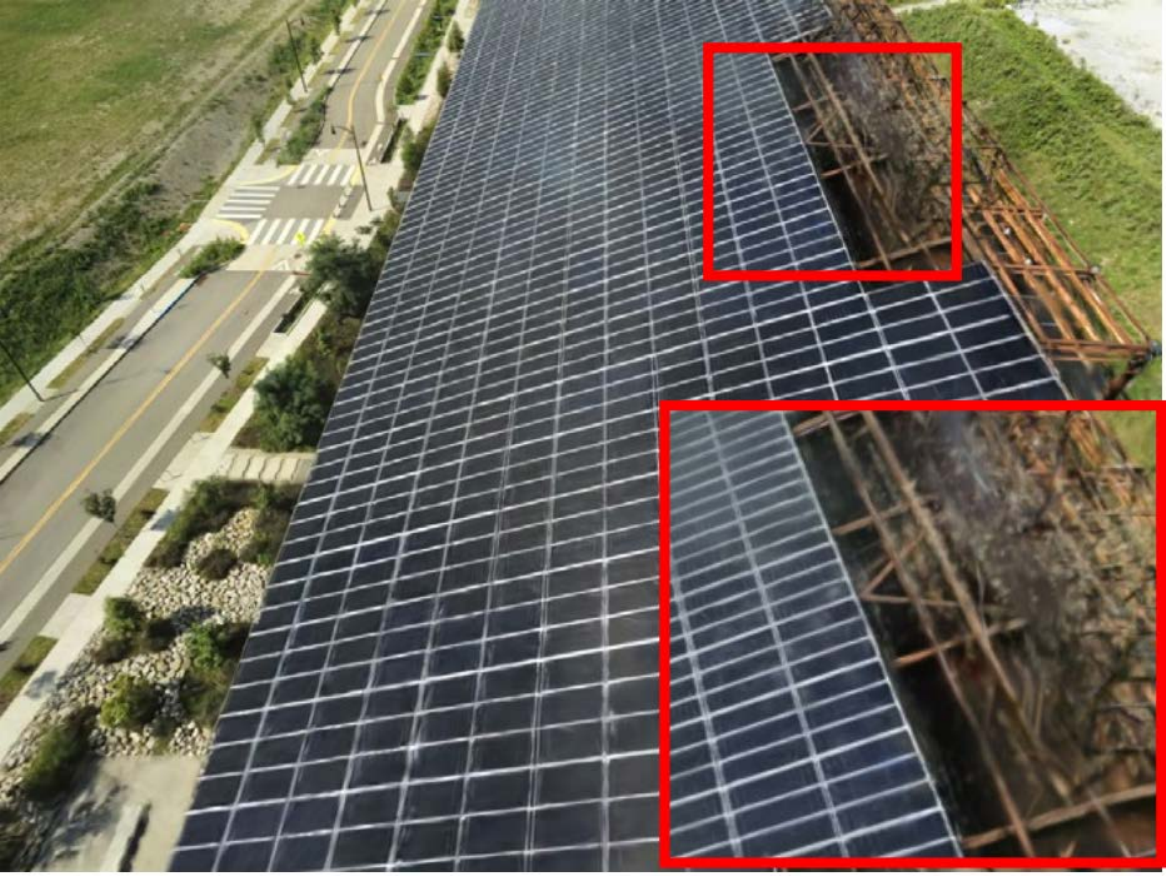}} & 
    \makecell{\includegraphics[width=.196\linewidth]{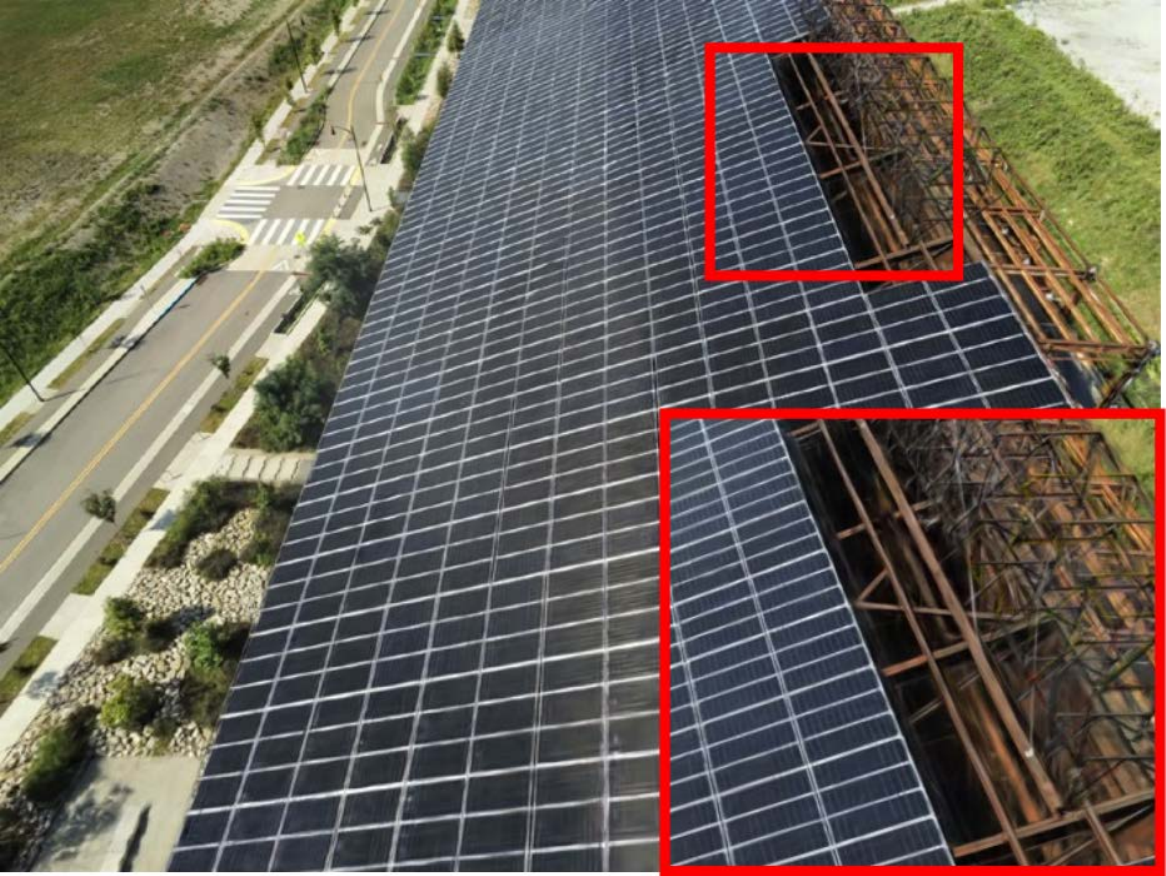}} & 
    \makecell{\includegraphics[width=.196\linewidth]{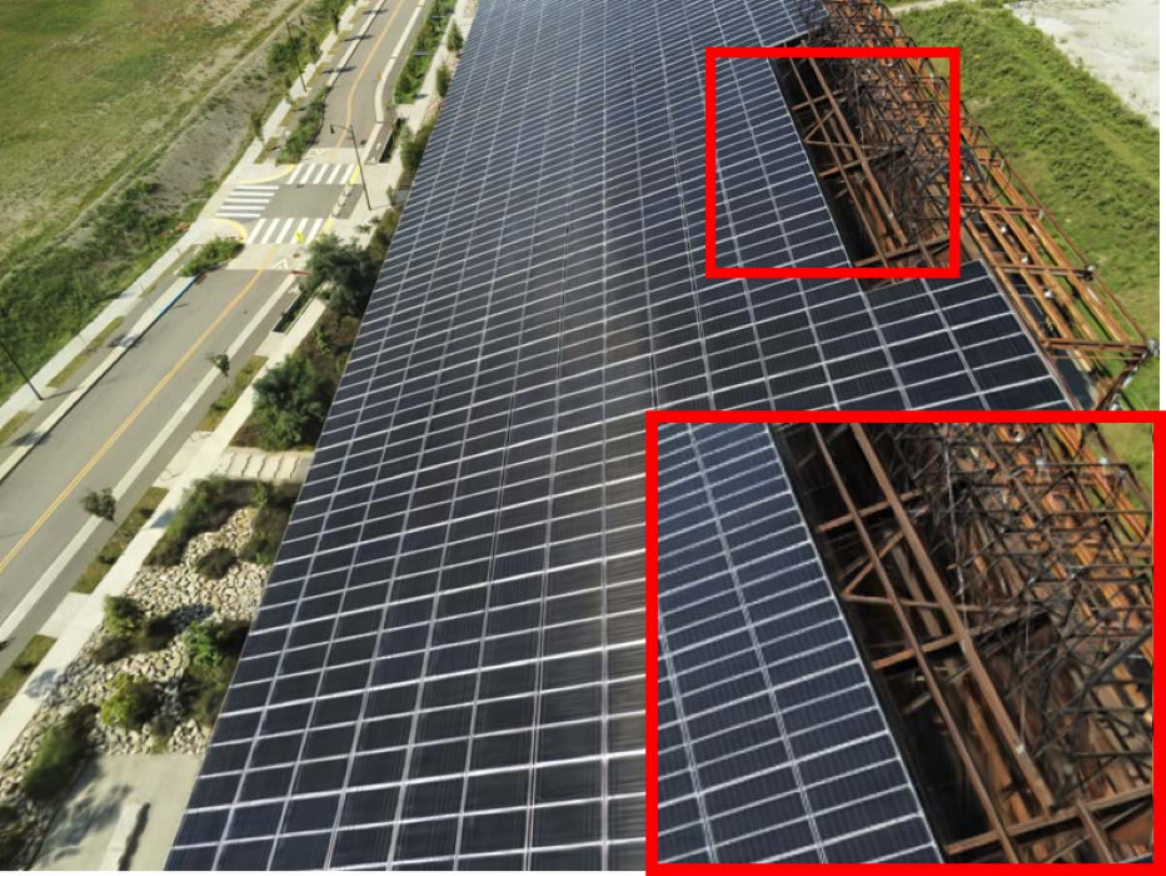}}\vspace{-2.5pt} \\
    \makecell{\includegraphics[width=.196\linewidth]{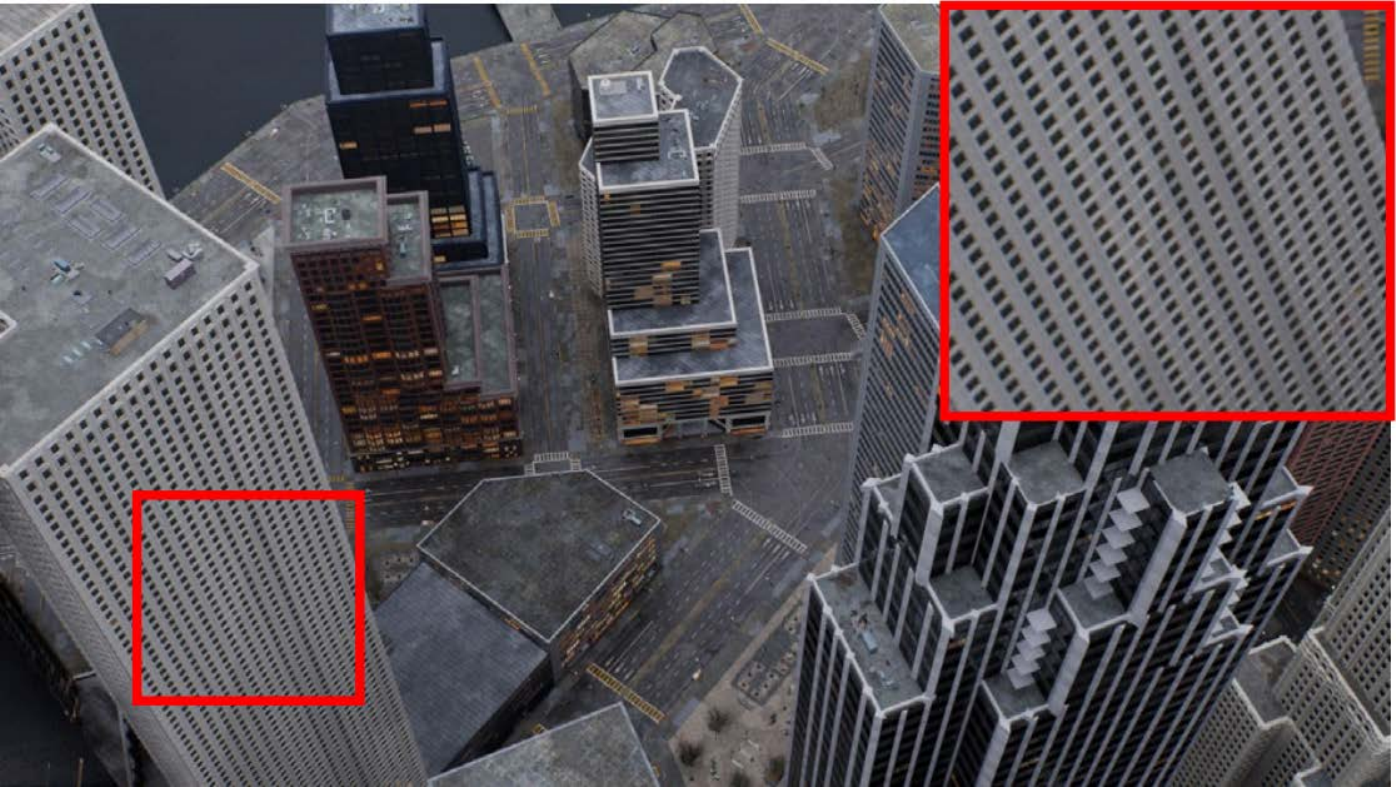}} & 
    \makecell{\includegraphics[width=.196\linewidth]{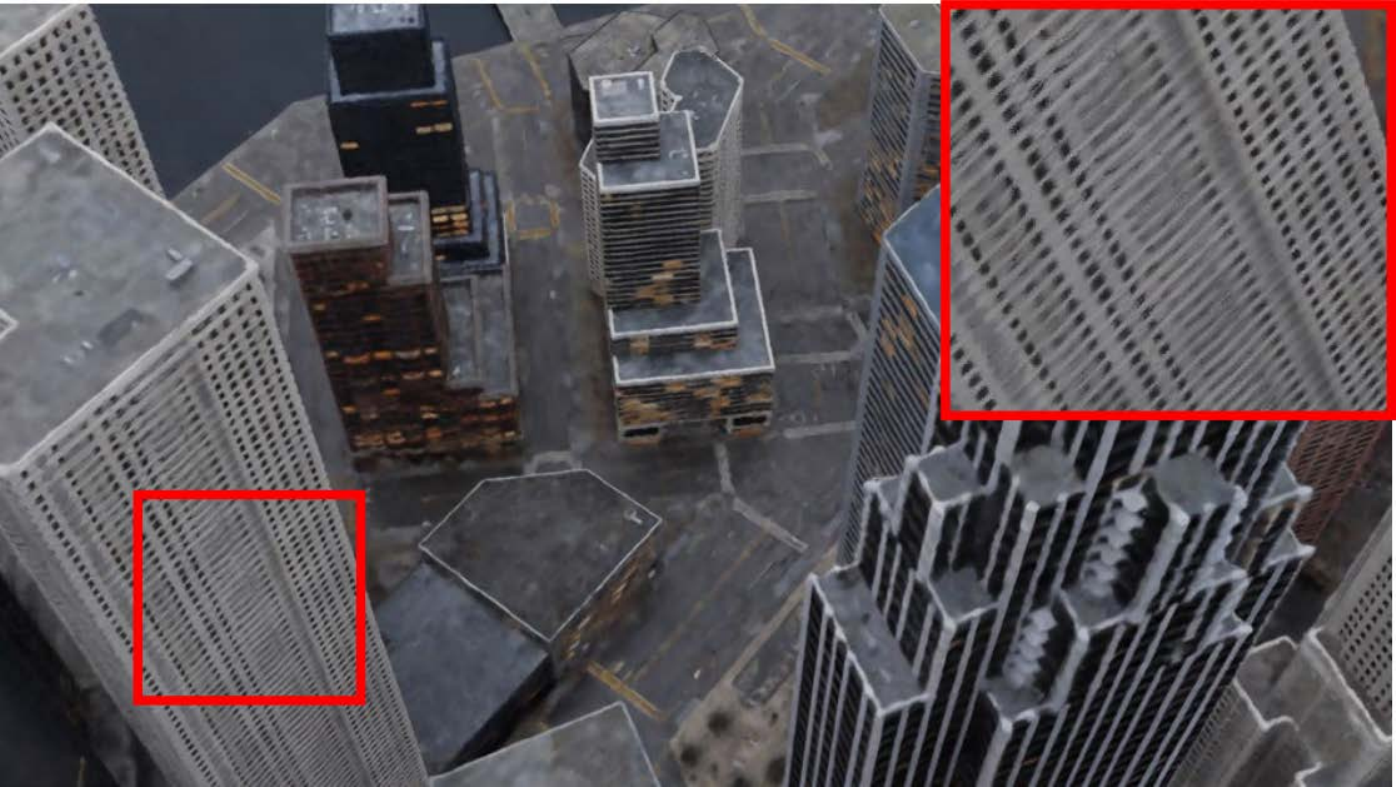}} &
    \makecell{\includegraphics[width=.196\linewidth]{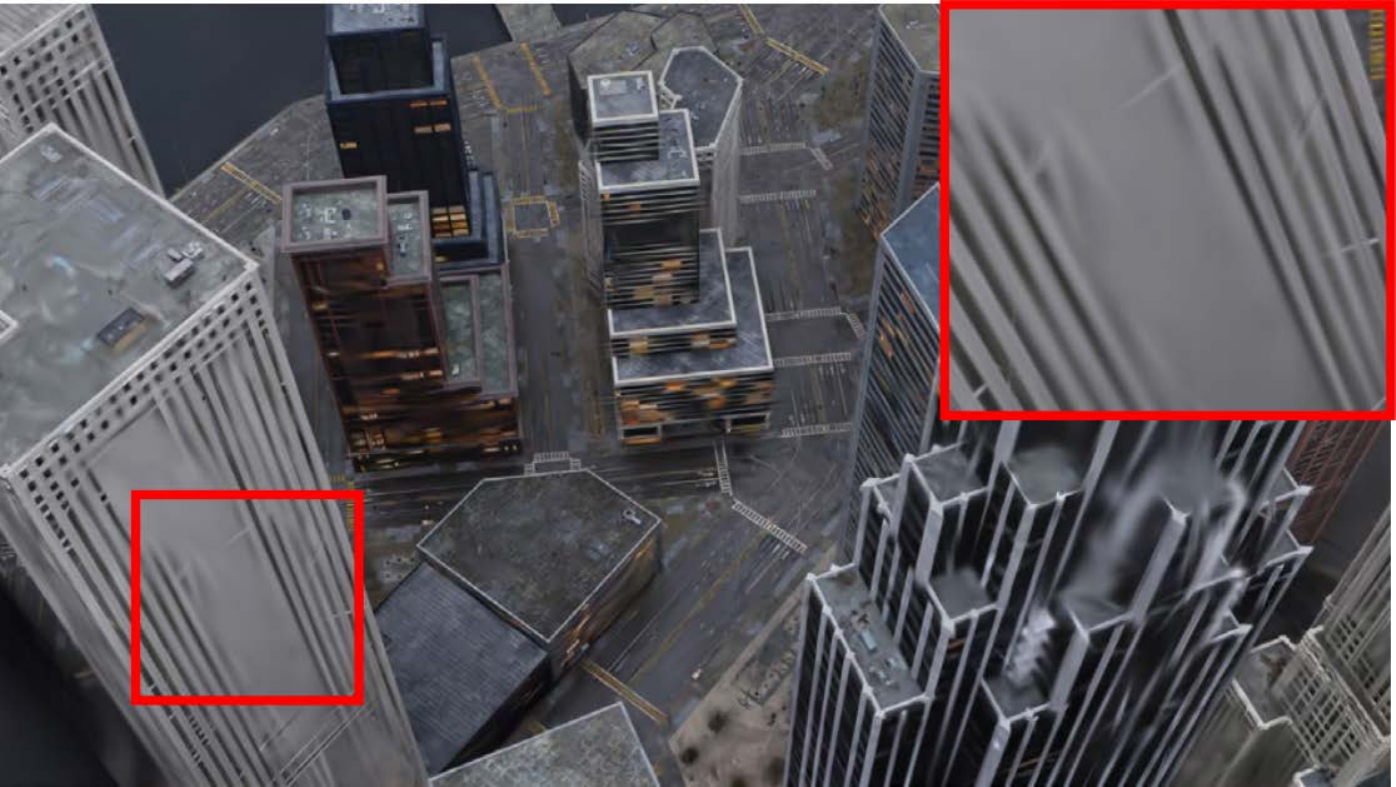}} & 
    \makecell{\includegraphics[width=.196\linewidth]{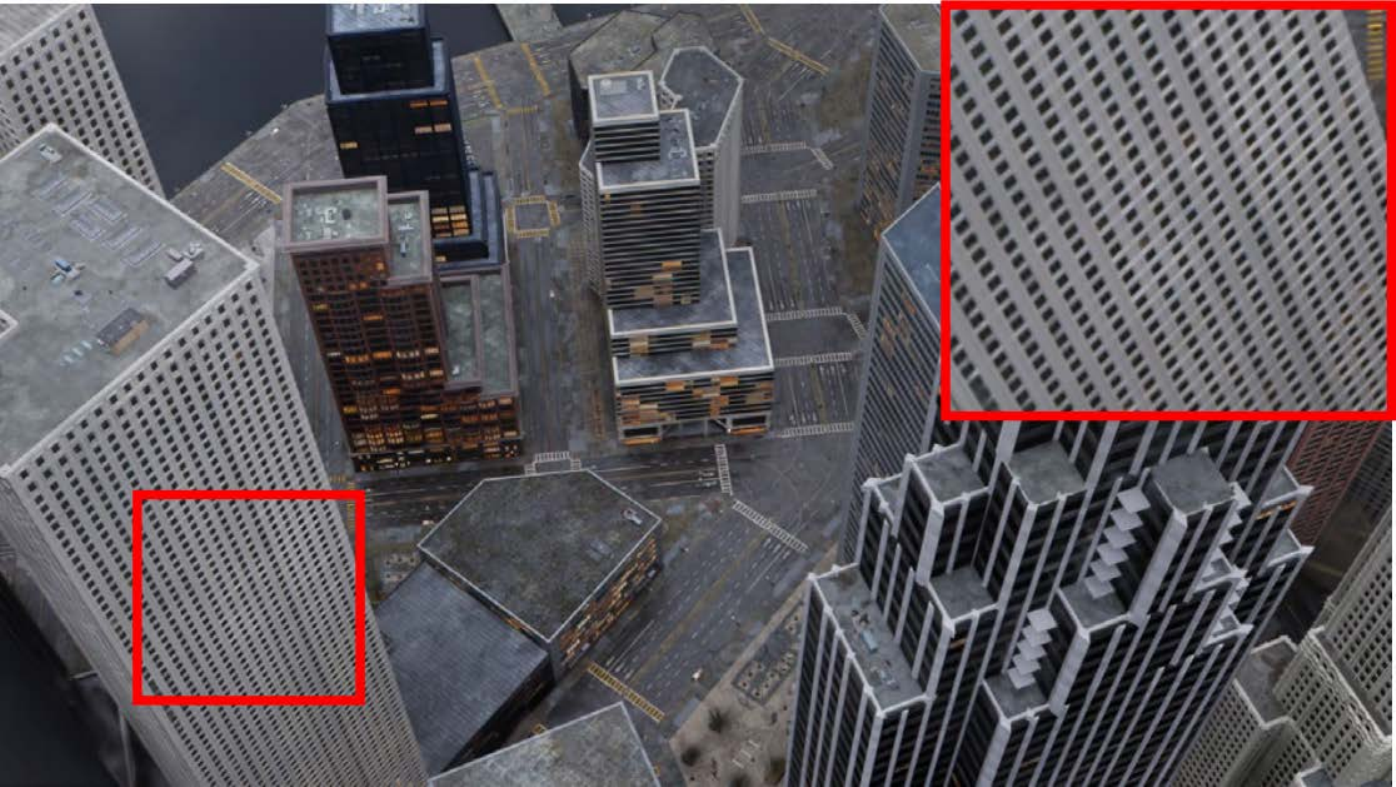}} & 
    \makecell{\includegraphics[width=.196\linewidth]{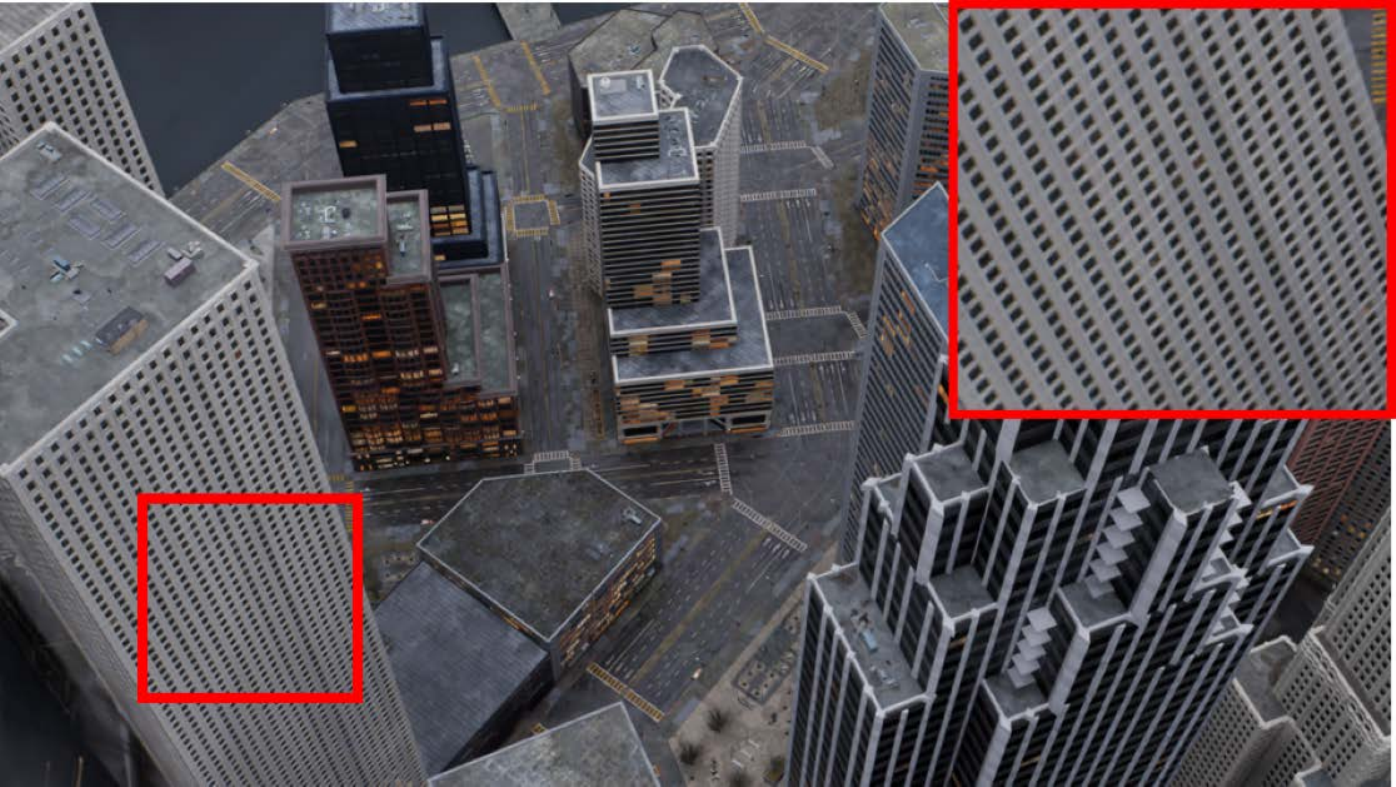}}\vspace{-2.5pt} \\
    \makecell{\includegraphics[width=.196\linewidth]{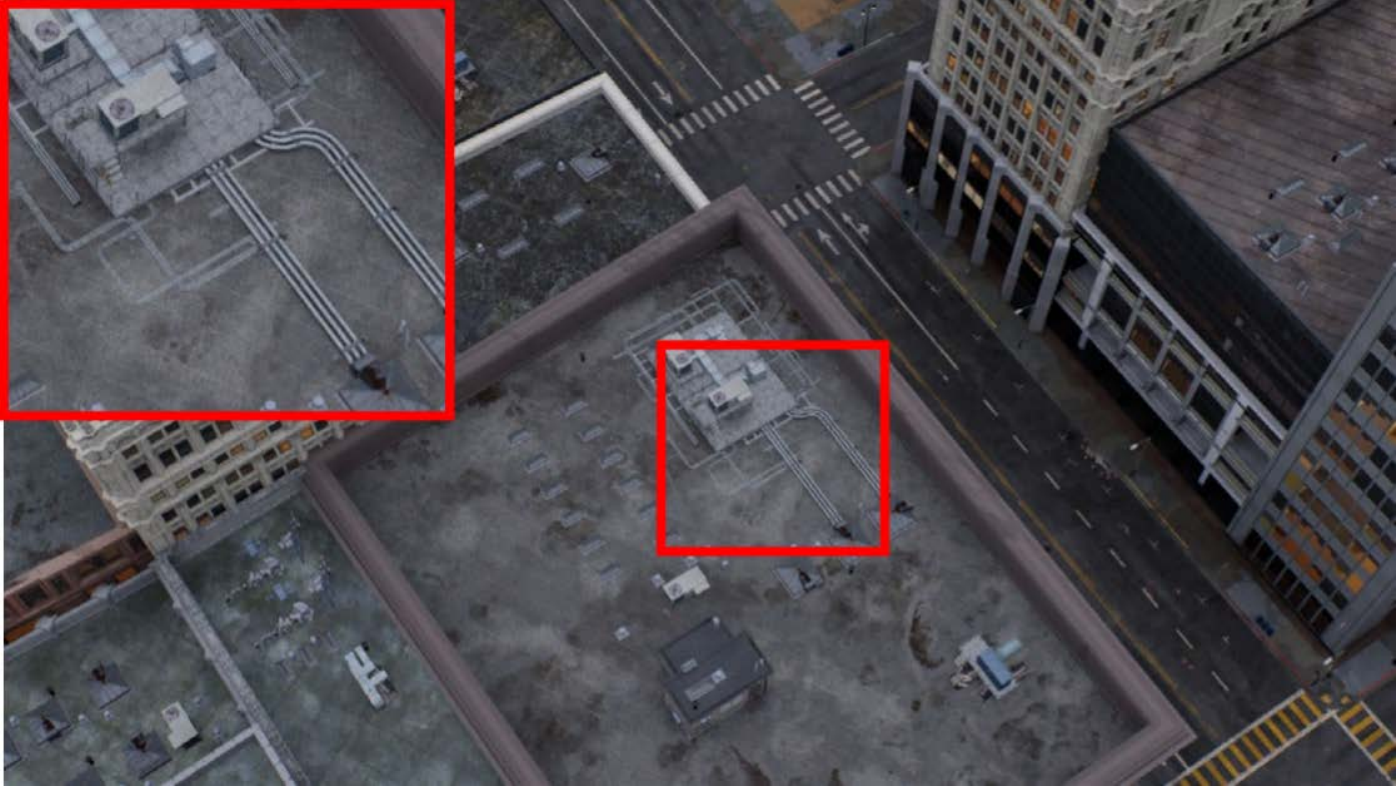}} & 
    \makecell{\includegraphics[width=.196\linewidth]{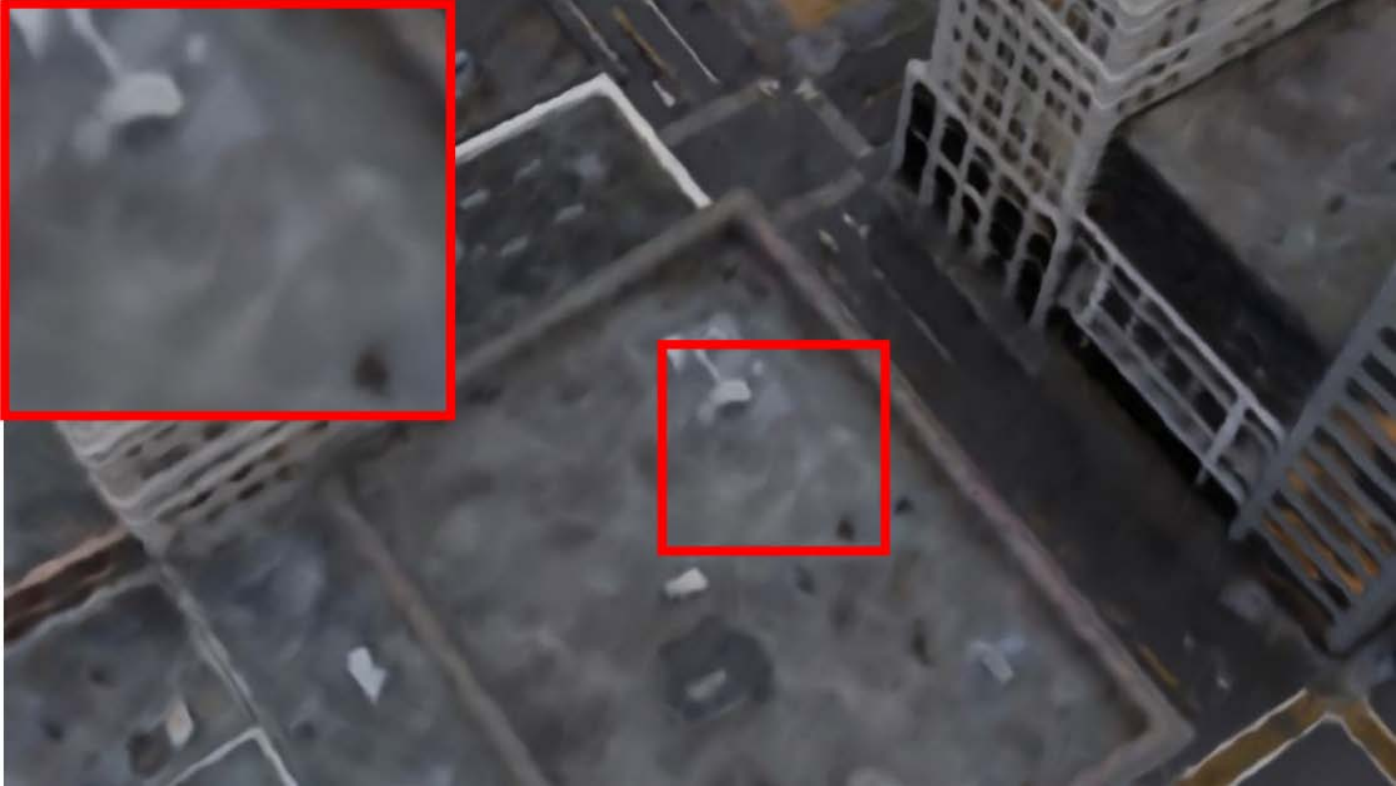}} &
    \makecell{\includegraphics[width=.196\linewidth]{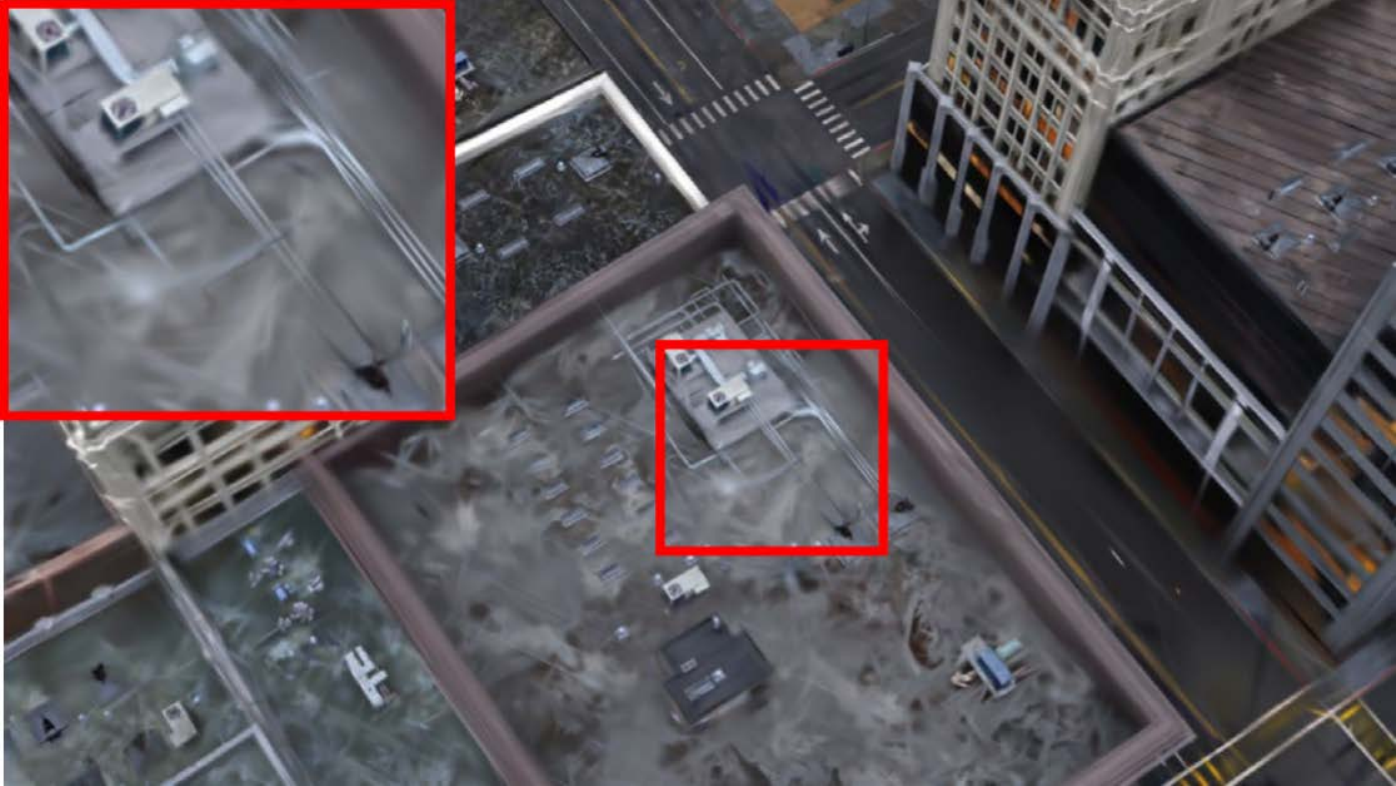}} & 
    \makecell{\includegraphics[width=.196\linewidth]{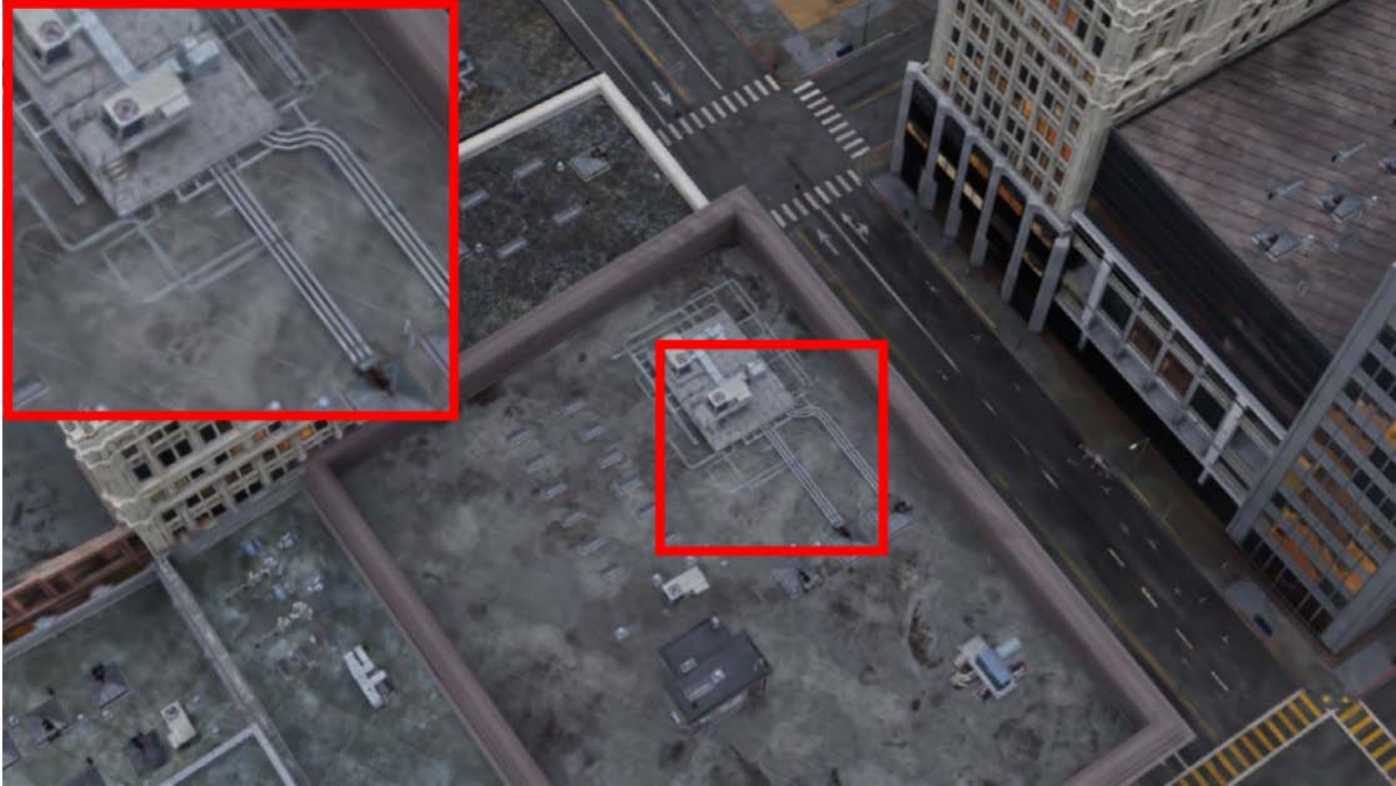}} & 
    \makecell{\includegraphics[width=.196\linewidth]{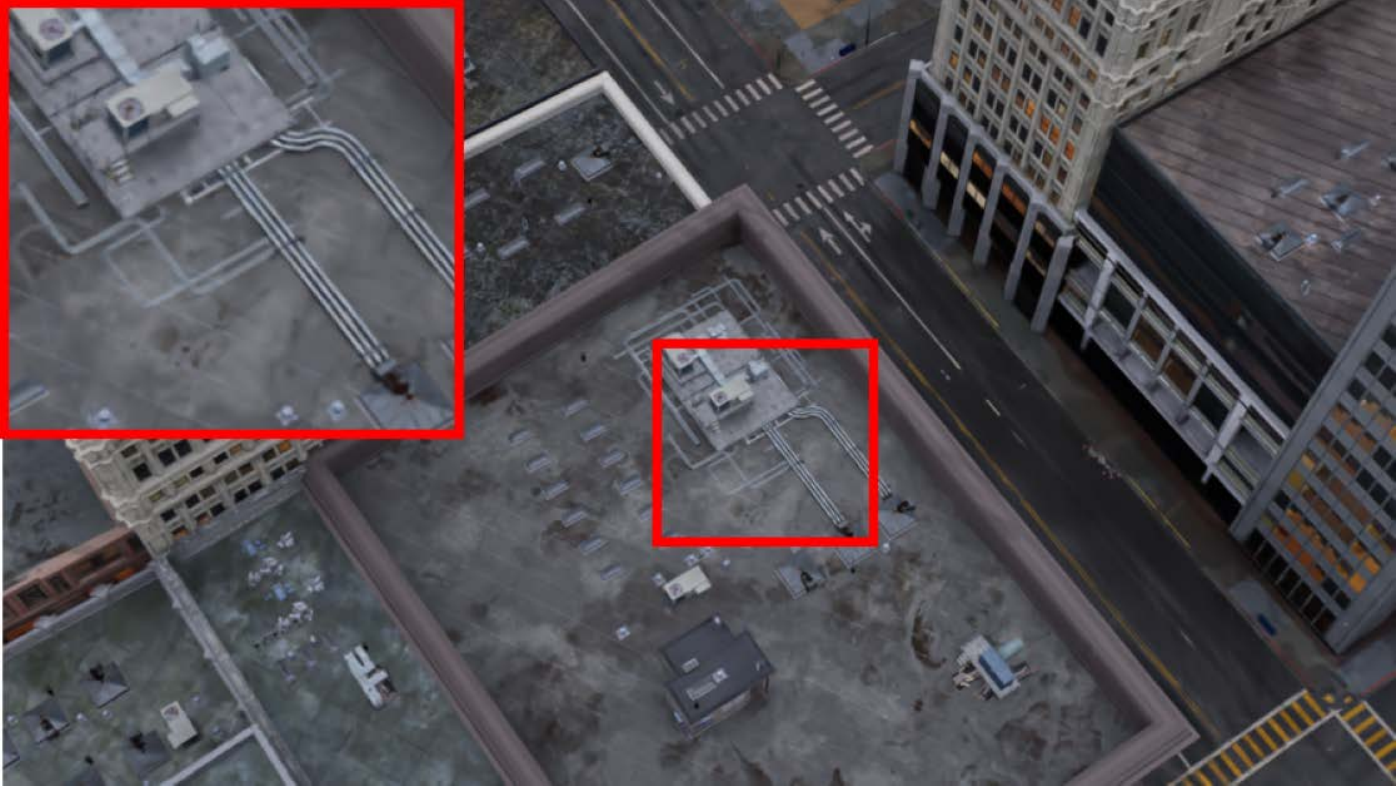}}
  \end{tabular}
\caption{\textbf{Qualitative Comparison with SOTA.} The first row represents the \textit{Rubble} scenario, the second row manifests the \textit{building} scenario, and the third and fourth rows showcase \textit{small\_city} scenes from the \textit{MatrixCity} dataset. The experiment demonstrates superior capability of our \FramewkName{ }framework in preserving color fidelity in rendered images, which is more closely resembling to the original images. Specifically, the region of interests are zoomed in with red box. \textbf{(Best viewed with zoom-in.)}}
    \label{fig:sota4data}\
\end{figure*}

\begin{table*}[h]
\begin{center}
\caption{\textbf{Quantitative comparison on four challenging datasets with SOTA large-scale reconstruction methods.} Symbol ‘-’ indicates that Mega-NeRf and Switch-NeRF are not evaluated on MatrixCity because of the difficulty in training them on different configuration. The \colorbox{lightred}{\bf red}, \colorbox{lightorg}{\bf orange} and  \colorbox{lightyellow}{\bf yellow} colors respectively denote the best, the second best, and the third best results.}
\begin{tabular}{l|ccc|ccc|ccc|ccc}
\hline
Scenes & \multicolumn{3}{c|}{Building}               & \multicolumn{3}{c|}{Rubble}& \multicolumn{3}{c|}{Residence}& \multicolumn{3}{c}{MatrixCity}    
\\ 
\hline 
Metrics  & \multicolumn{1}{c}{SSIM$\uparrow$} & \multicolumn{1}{c}{PSNR$\uparrow$} & \multicolumn{1}{c|}{LPIPS$\downarrow$} & \multicolumn{1}{c}{SSIM$\uparrow$} & \multicolumn{1}{c}{PSNR$\uparrow$} & \multicolumn{1}{c|}{LPIPS$\downarrow$} & \multicolumn{1}{c}{SSIM$\uparrow$} & \multicolumn{1}{c}{PSNR$\uparrow$} & {LPIPS$\downarrow$} & \multicolumn{1}{l}{SSIM$\uparrow$} & \multicolumn{1}{l}{PSNR$\uparrow$} & {LPIPS$\downarrow$} 
\\
\hline 
Mega-NeRF \cite{turki2022mega}  & 0.550 & 20.85   & 0.499   & 0.561    & 24.09    & 0.509   & 0.625  & 22.10   & 0.481  & \multicolumn{1}{c}{-}  & \multicolumn{1}{c}{-}     & \multicolumn{1}{c}{-} 
\\
Switch-NeRF\cite{zhenxing2022switch}  & 0.577  & {\scd21.67}    & {\trd0.480}   & 0.569    & 23.50    & 0.501 & 0.656  & {\scd22.60}  & 0.460  & \multicolumn{1}{c}{-}   & \multicolumn{1}{c}{-}     & \multicolumn{1}{c}{-} 
\\
GP-NeRF \cite{xu2023grid}   & 0.570  & 21.10            & 0.489    & 0.565  & 24.32    & 0.489     & 0.659            & {\trd22.29}  & 0.450  & 0.610   & 23.60    & 0.392   
\\   
3DGS \cite{kerbl20233d}        & {\trd0.731}    & 20.50  & {\scd0.307}   & {\trd0.790}    & {\trd25.69}   & {\trd0.281}   & {\trd0.800}  & 22.10   & {\trd0.229}  & {\trd0.740}   & {\trd23.71}    & {\trd0.390}       
\\
CityGaussian \cite{liu2024citygaussian} & {\scd0.780}  & {\trd21.61}  & {\scd0.307}   & {\scd0.821}    & {\scd27.00}   & {\scd0.219}  & {\scd0.820}   & 21.19  & {\scd0.220}  & {\scd0.868}      & {\scd27.53}  & {\scd0.200}  
\\
\hline
Ours  & {\fst0.818}  & {\fst25.32}       & {\fst0.227}  & {\fst0.866}     & {\fst29.19}        & {\fst0.198}    & {\fst0.839}      & {\fst22.72}  & {\fst0.214} & {\fst0.897}   & {\fst28.73}   & {\fst0.194}  \\ \hline
\end{tabular}
\label{Tab:compare}
\end{center}
\end{table*}

Our approach is compared to Mega-NeRF \cite{turki2022mega}, Switch-NeRF \cite{zhenxing2022switch}, GP-NeRF \cite{zhang2023efficient},  3DGS \cite{kerbl20233d}, and CityGS \cite{liu2024citygaussian}. First, we stop densification at 15k iterations as 3DGS \cite{kerbl20233d}. Given that most of the data sets consist of a significantly larger number of input images than the data sets used in 3DGS \cite{kerbl20233d}, we adjusted the total number of training iterations to 60k. The default camera visibility is set to 0.25.

\subsection{Performance of Novel view Synthesis}
\subsubsection{\textbf{Comparison with SOTA}}
As demonstrated in Table \ref{Tab:compare}, our method outperforms the state-of-the-art (SOTA) methods in terms of SSIM, PSNR, and LPIPS metrics for all four scenes (\textit{Building}, \textit{Rubble}, \textit{Residence}, and \textit{MatrixCity}. Here, \textit{small\_city} is selected from the \textit{MatrixCity} dataset.).
The qualitative results presented in Fig. \ref{fig:sota4data} also validate the high fidelity of our rendering results. As shown in Fig. \ref{fig:sota4data}, our rendering results achieve more realistic results, which is much closer to Ground Truth in the aspects of lighting and color. Specifically, in the building scenario, our renderings better preserve the detail of sunlight reflection on solar panels, which validates the effectiveness of our ray-Gaussian-intersection rendering, density control strategy, and the color decoupling module based on KAN and CNN.

\subsubsection{\textbf{Experiments on Self-collected Data}}
\begin{figure}[H]
  \centering
  \setlength{\tabcolsep}{1pt}

  \begin{tabular}{ccc}
   \subcaptionbox{Ground Truth}{\includegraphics[width=.329\linewidth]{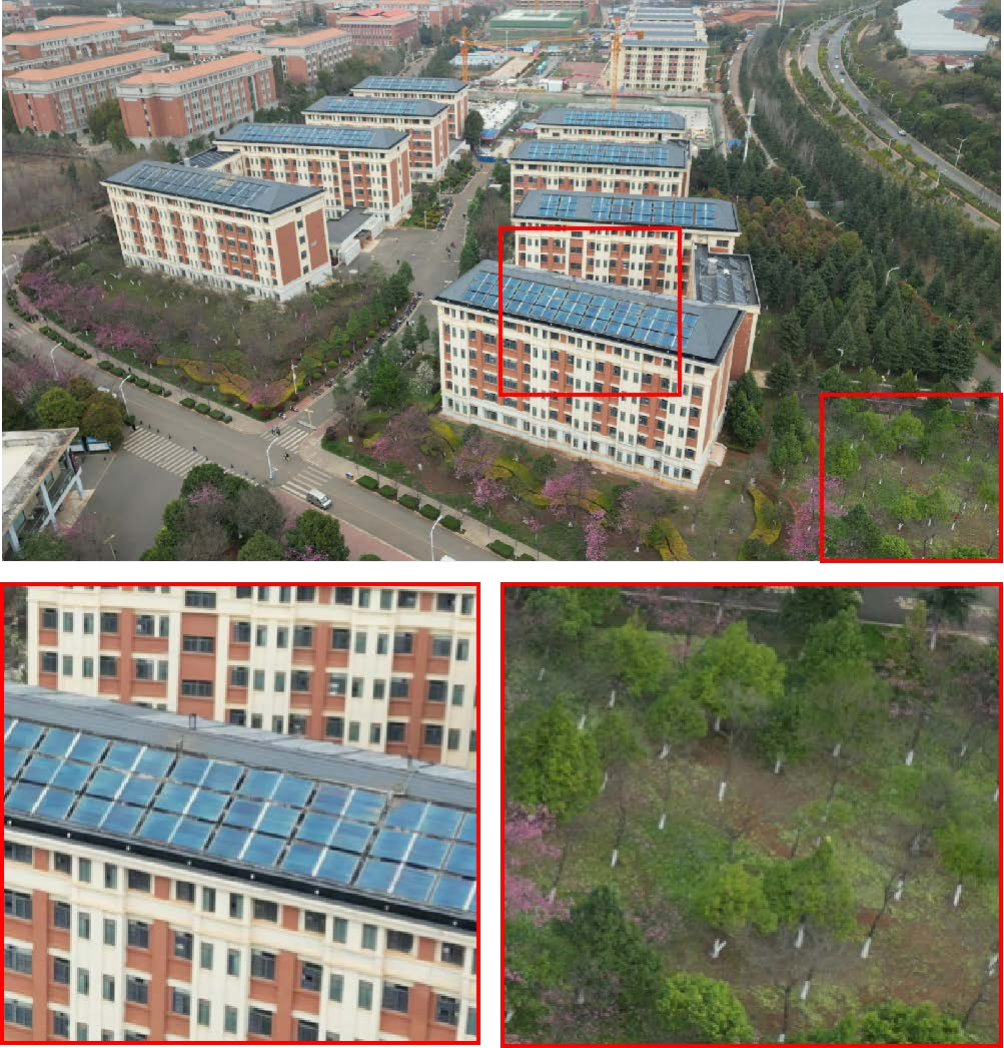}} & 
    \subcaptionbox{3DGS \cite{kerbl20233d}}{\includegraphics[width=.329\linewidth]{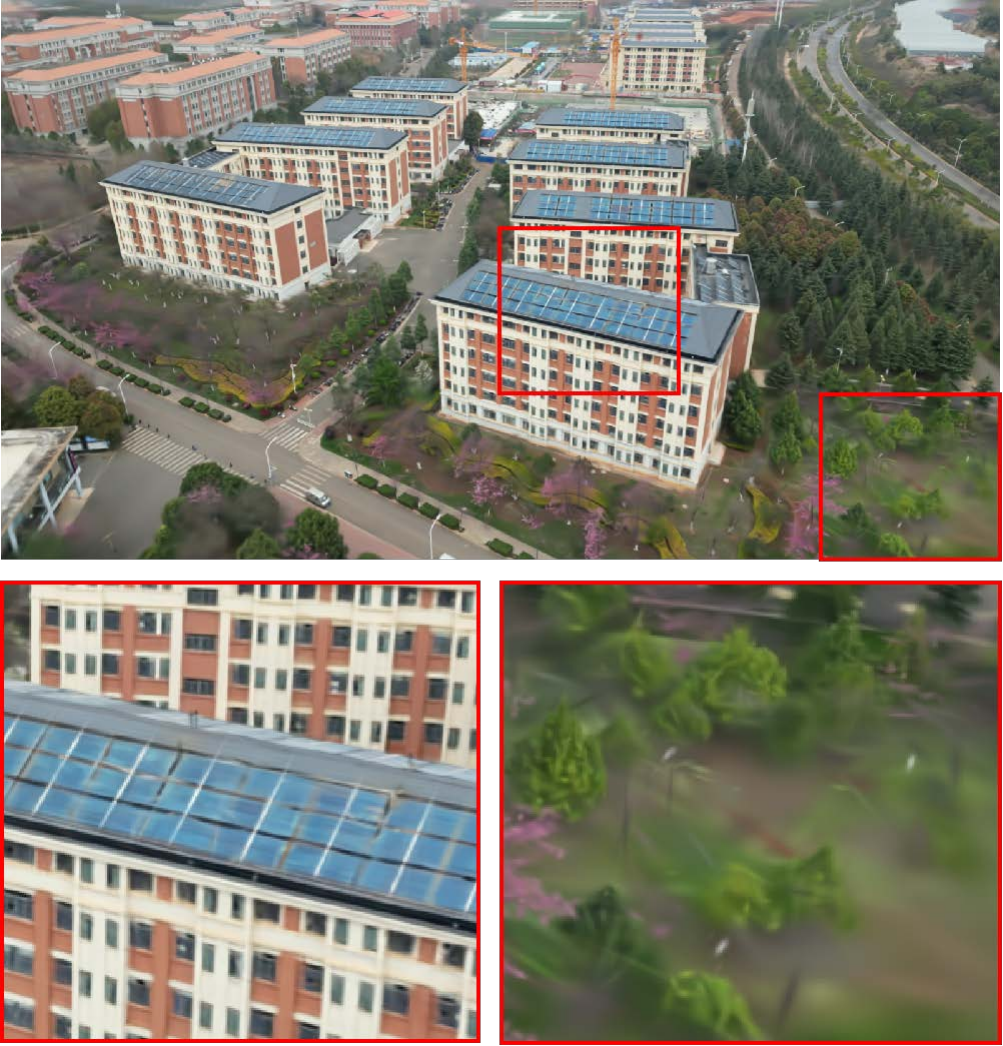}} &
    \subcaptionbox{Ours}{\includegraphics[width=.329\linewidth]{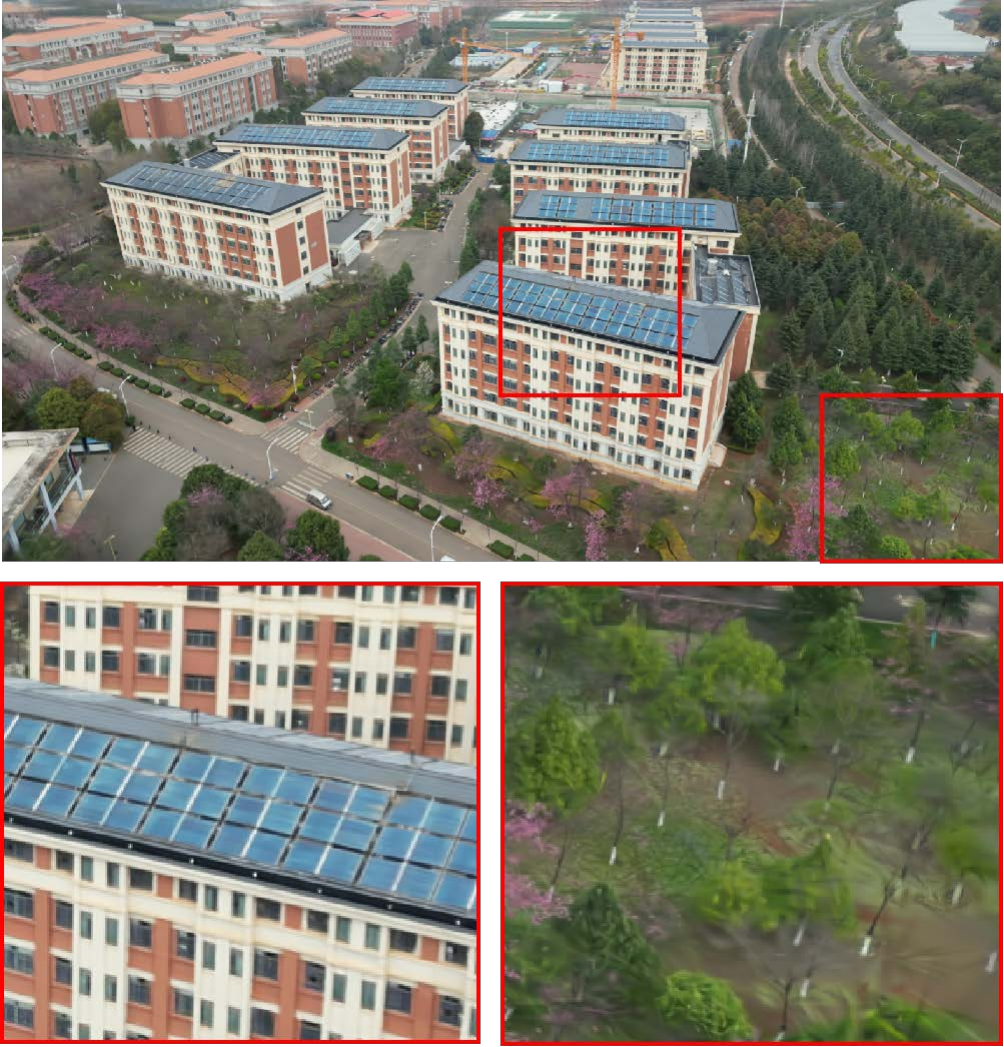}} 
  \end{tabular}
  \caption{\textbf{Comparison of our method with 3DGS on Self-collected data.} Fig. \ref{fig:ablation_self}.a corresponds to the original image obtained from \textit{Campus-YNU} scenario Fig. \ref{fig:ablation_self}.b illustrates the image rendered using 3DGS, where the solar panels and the trees exhibit a degree of blurriness. Fig. \ref{fig:ablation_self}.c demonstrates the image rendered with our proposed method, showing a decent enhancement in the clarity of the solar panels and the trees. \textbf{(Best viewed with zoom-in.)}}
  \label{fig:ablation_self}
        \vspace{-0.3cm}
\end{figure}

To validate the effectiveness and generalization capability of our framework in large-scale scenarios, we employ a DJI drone (Mini 3 Pro) to fly over a $1 km \times 1 km$ area, captured a dataset comprising 1,600 images at a resolution of $3768 \times 2118$ pixels. This scene was selected in our experiment for its intricate details, including solar panels, window arrays, and construction sites. Comparative experiments are conducted between 3DGS and our \FramewkName{}. As shown in Table \ref{tab:campsYNU}, the results of the experiment demonstrate an obvious improvement of our method over 3DGS considering the rendering quality. In Fig. \ref{fig:ablation_self}, we can observe that our method makes solar panels and grasslands more precise compared to 3DGS.

\begin{table}[H]
\caption{\textbf{Experiments on Self-collected Campus-YNU Dataset.} We validate the effectiveness of our method compared to vanilla 3DGS on our self-collected data.}
\label{tab:campsYNU}
\centering
\begin{tabular}{lccc}
\hline
\multicolumn{1}{l|}{Metrics} & \multicolumn{1}{c}{SSIM} & \multicolumn{1}{c}{PSNR} & \multicolumn{1}{c}{LPIPS} \\ \hline
\multicolumn{1}{l|}{3DGS \cite{kerbl20233d} } & 0.831 & 28.95 & 0.241 \\
\multicolumn{1}{l|}{Ours} &\fst 0.896 & \fst31.58 & \fst0.151 \\ \hline
\end{tabular}
\vspace{-0.4cm}
\end{table}

\subsection{Ablation Study}
We conduct the ablation study on the Rubble scenario and our self-collected data to evaluate the different proposed techniques of our method. We randomly select 95\% of the images from the Rubble dataset as the training set and 5\% as testing set, and the ratio is kept same to our Campus-YNU dataset.

\begin{table}[H]
\caption{\textbf{Quantitative Results of the Ablation Study.} The \colorbox{lightred}{\bf red}, \colorbox{lightorg}{\bf orange} and  \colorbox{lightyellow}{\bf yellow} colors respectively denote the best, the second best, and the third best results. Full stands for the configuration with CNN + KAN + ViS R2 + Full Loss.}
\label{tab:ablation_std}
\begin{footnotesize} 
\begin{tabular}{l|ccc|ccc}
\hline
Scene & \multicolumn{1}{l}{} & Rubble & \multicolumn{1}{l|}{} & \multicolumn{3}{c}{Campus-YNU (Self-collect.)} \\ \hline
Metrics & SSIM & PNSR & LPIPS & SSIM & PNSR & LPIPS \\ \hline
Vis R0 & \trd0.853 & \trd29.18 & 0.592 & \trd0.889 & 30.56 & \trd0.161 \\
Vis R1 & \scd0.857 & \fst29.33 & 0.229 & \scd0.895 & \scd30.90 & \scd0.152 \\
CNN & 0.835 & 28.81 & \trd0.226 & 0.885 & 29.10 & 0.178 \\
 $\mathcal{L}_{c} $ Only & 0.795 & 27.29 & \scd0.205 & 0.887 & \trd30.70 & 0.164 \\ \hline
Full & \fst0.854 & \scd29.09 & \fst0.201 & \fst0.896 & \fst31.58 & \fst0.151 \\ \hline
\end{tabular}
\end{footnotesize}
\end{table}

\subsubsection{\textbf{Camera Visibility Calculation}}
As shown in the Table \ref{tab:ablation_std}, we conduct an investigation into camera visibility within a \textit{Rubble} scenario. Specifically, we establish three levels of camera visibility, designated as Vis R0, Vis R1, and Vis R2, with settings of 0, 0.50, and 0.25, respectively. Throughout the experimental process, we use a full loss function and a color decoupling module integrated with CNN and KAN \cite{liu2024kan}. Subsequently, our approach employs the proposed camera visibility technique to test all these levels of visibility. As demonstrated in Fig. \ref{fig:AblatStudy}.d, Fig. \ref{fig:AblatStudy}.e, and Fig. \ref{fig:AblatStudy}.f, the camera visibility is found to be helpful in enhancing the quality of the rendering.

\subsubsection{\textbf{Loss}}
Our loss function, which is composed of depth distortion loss, normal consistency loss, and RGB loss derived from 3DGS \cite{kerbl20233d}, can better enhance the rendering quality of images compared to using only the RGB loss from 3DGS \cite{kerbl20233d}. As illustrated in Fig. \ref{fig:AblatStudy}.h and Fig. \ref{fig:AblatStudy}.i, the rendered text is obviously clearer after using our loss function.

\subsubsection{\textbf{Decoupled Color Model}}
\begin{figure}[ht!]
      \vspace{-0.2cm}

  \centering
  \setlength{\tabcolsep}{1pt}
  \begin{tabular}{ccc}
    {\scriptsize (a) Ground Truth 1} & {\scriptsize (b) CNN} & {\scriptsize (c) KAN + CNN} \\
    \includegraphics[width=0.327\linewidth]{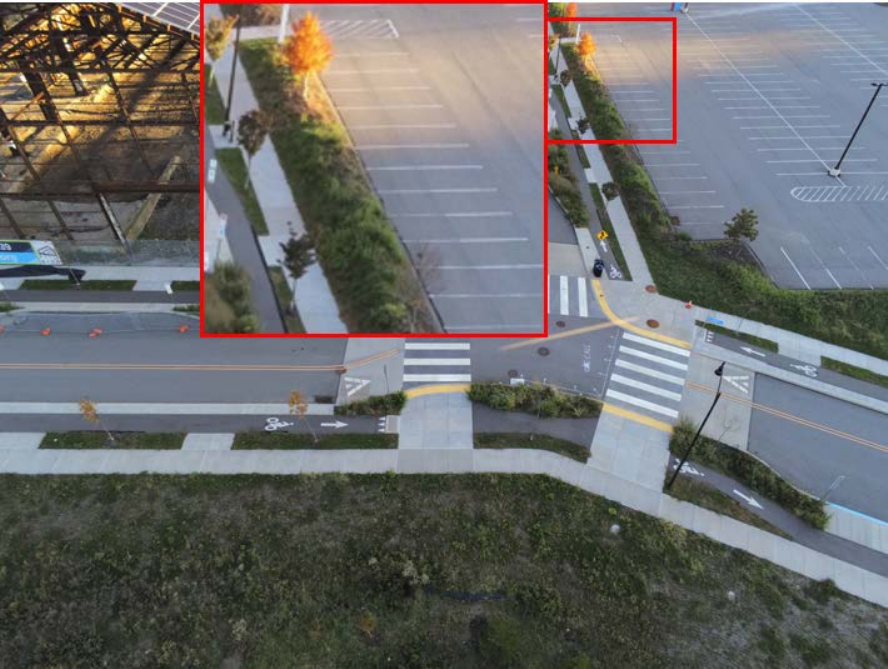} & 
    \includegraphics[width=0.327\linewidth]{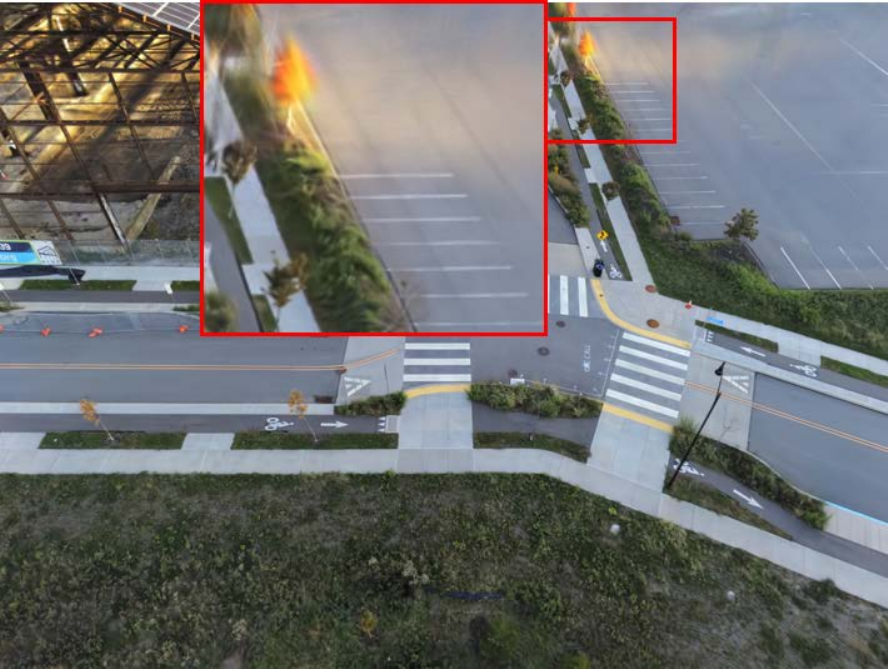} & 
    \includegraphics[width=0.327\linewidth]{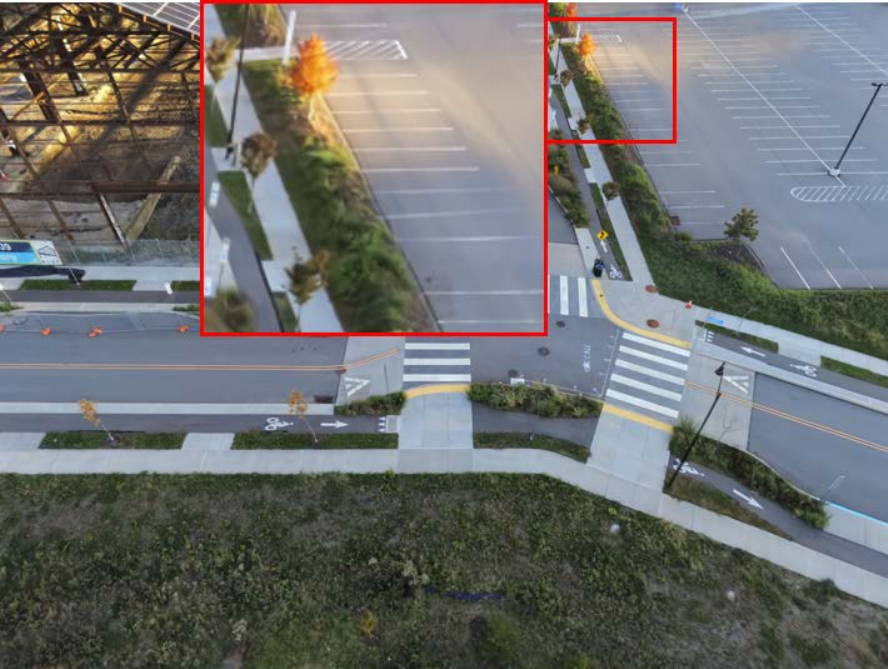} \\
    
    {\scriptsize (d) Vis R0} & {\scriptsize (e) Vis R1} & {\scriptsize (f) Vis R2} \\
    
    \includegraphics[width=0.327\linewidth]{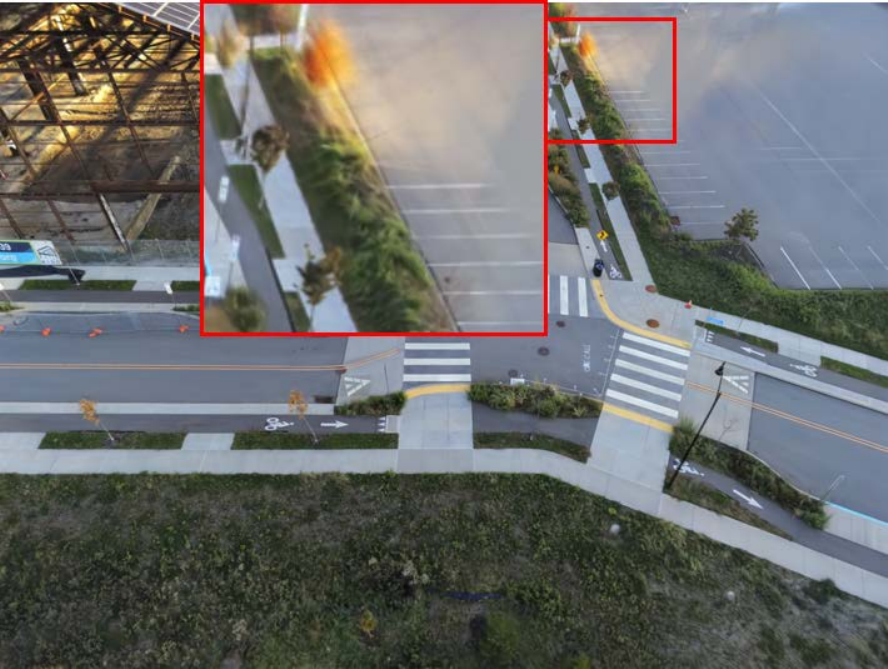} & 
    \includegraphics[width=0.327\linewidth]{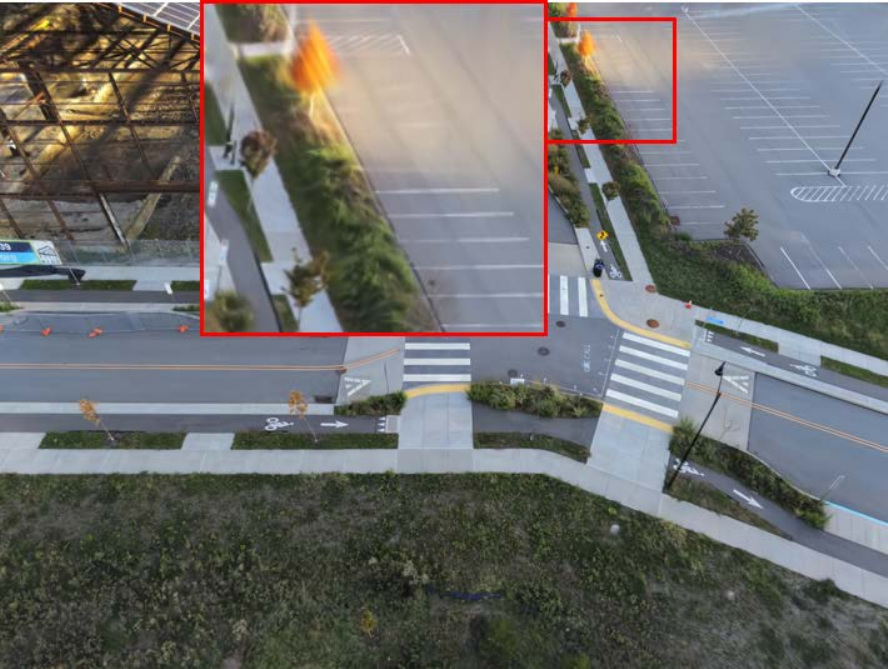} & 
    \includegraphics[width=0.327\linewidth]{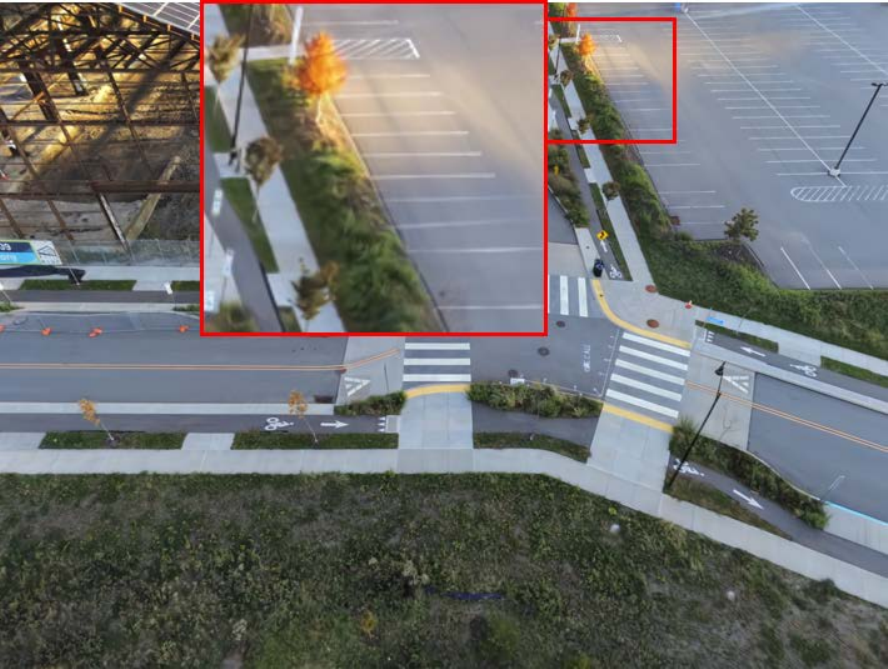} \\
   
    {\scriptsize (g) Ground Truth 2} & {\scriptsize (h) $\mathcal{ L}_{c} $ 
     Only} & {\scriptsize (i) Full Loss} \\

    \includegraphics[width=0.327\linewidth]{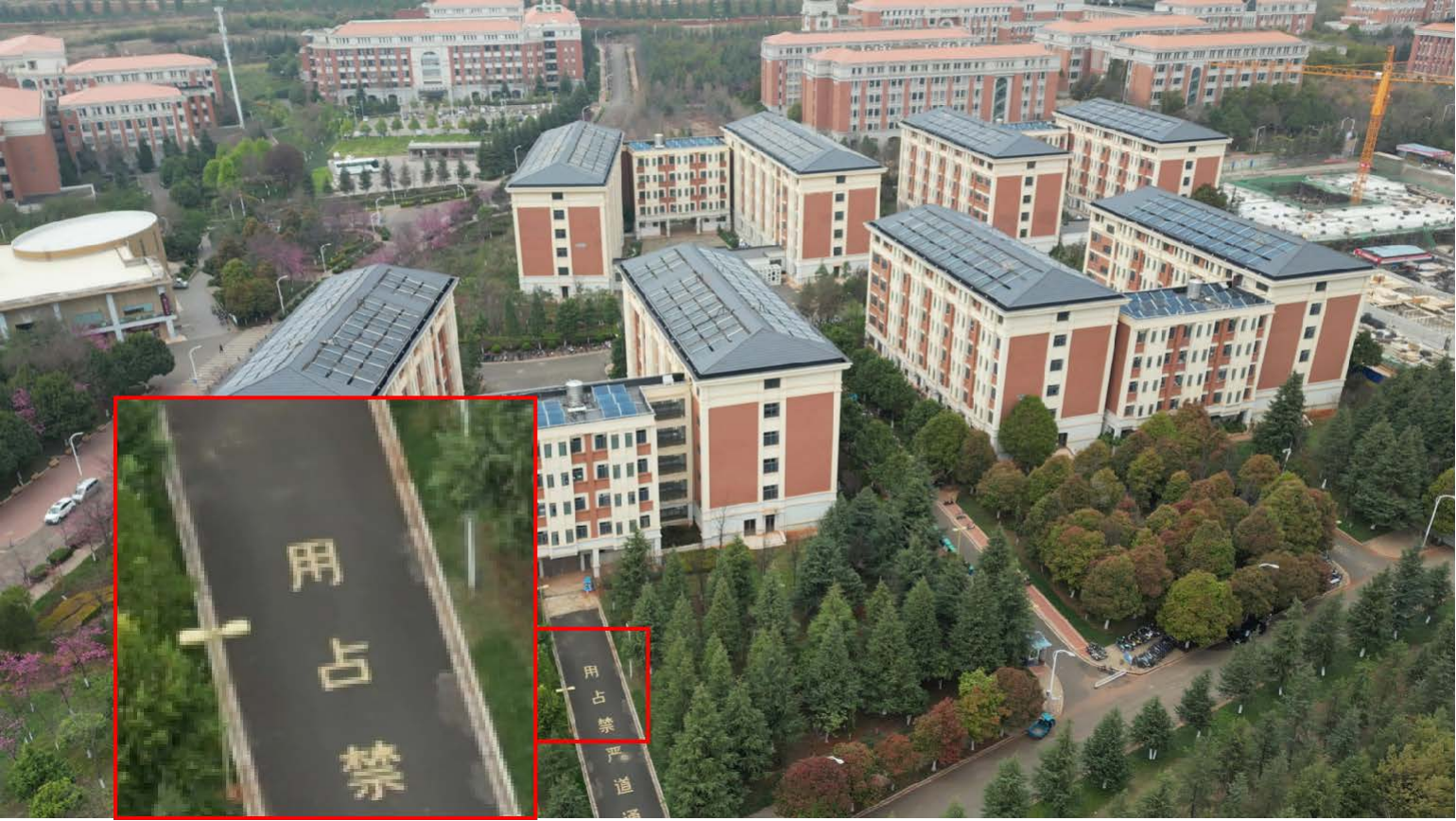} & 
    \includegraphics[width=0.327\linewidth]{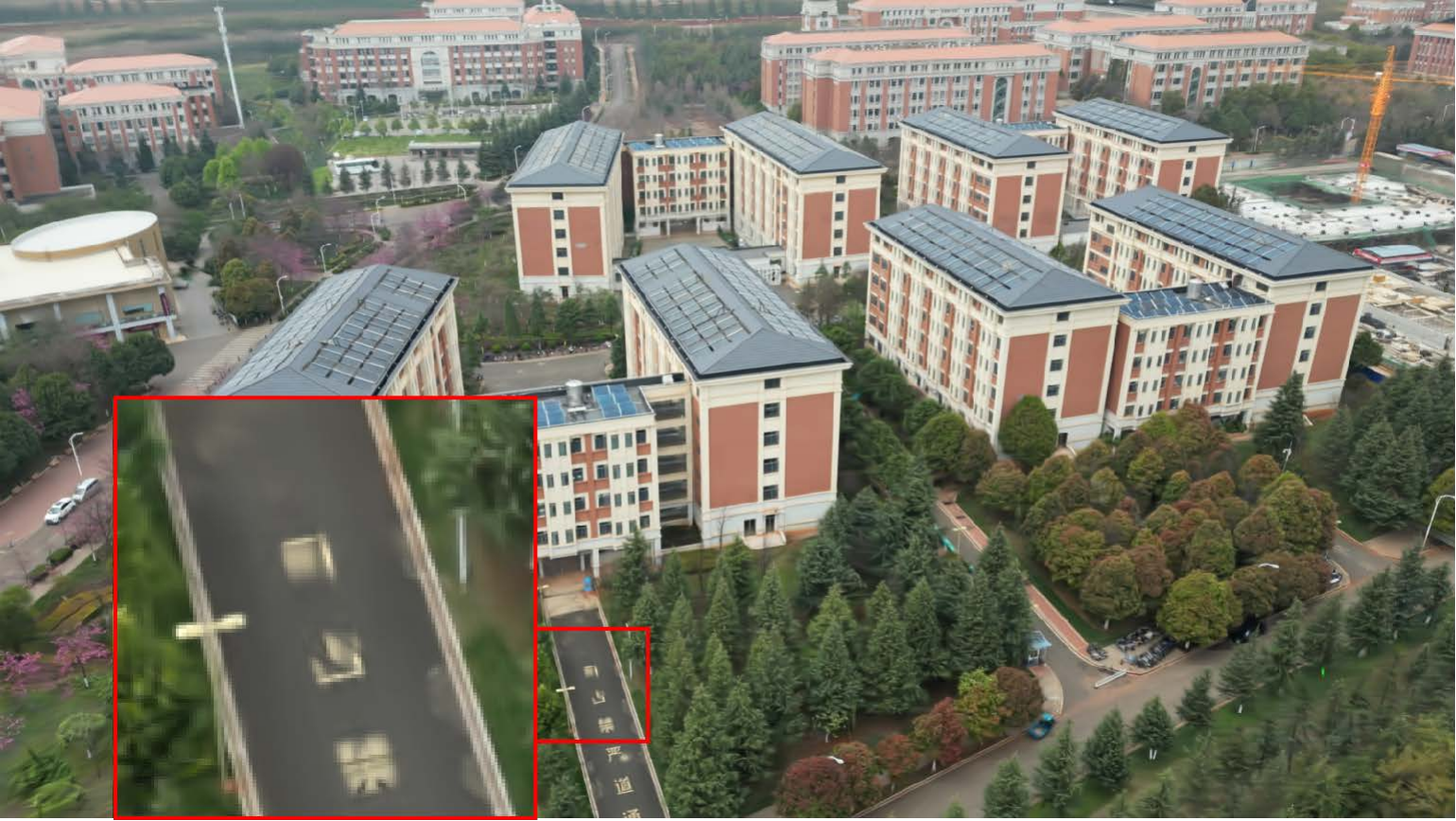} & 
    \includegraphics[width=0.327\linewidth]{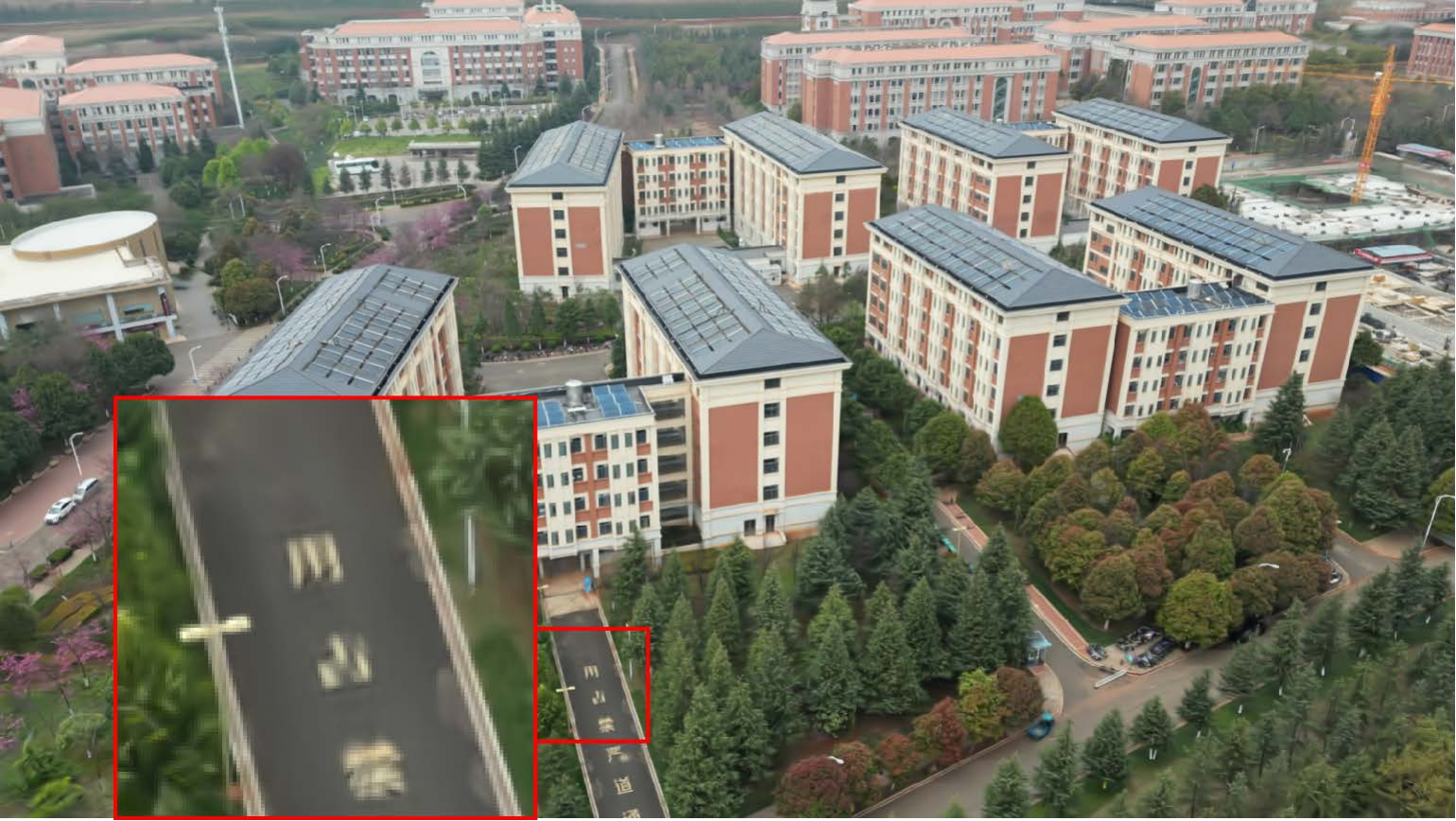} \\
    
  \end{tabular}
 \caption{\textbf{Qualitative Results of the Ablation Study.} Ground Truth 1 represents the original image from the \textit{Rubble} scenario.
Ground Truth 2 corresponds to the original image obtained from the \textit{Campus-YNU} scenario. \textbf{ (Best viewed with zoom-in.)}}
  \label{fig:AblatStudy}
      \vspace{-0.4cm}

\end{figure}

Our color decoupling module, which employs a network combining KAN \cite{liu2024kan} and CNN, has achieved superior results in reducing color variations in rendered images. As illustrated in Fig. \ref{fig:AblatStudy}.b and Fig. \ref{fig:AblatStudy}.c, compared to a color decoupling module composed solely of CNN, our approach can more effectively learn consistent geometric shapes and colors from training images with varying appearances. During the ablation experiments of the color decoupling module, we utilized a camera visibility of 0.25 and employed a full loss function.

%% file: 05_CONCLUS.tex
\section{CONCLUSION}\label{sec:conclusion}
In this work, we introduce \FramewkName{}, a high-fidelity reconstruction and rendering method for large-scale scenes based on 3D Gaussian splatting. We employ a ray-Gaussian-intersection volume rendering and a density control strategy for large-scale reconstruction, a color decoupling module that combines KAN and CNN, a data partitioning method based on coordinates and camera visibility, and depth-normal consistency. We have achieved state-of-the-art rendering fidelity in mainstream benchmark tests and excellent rendering fidelity in our self-collected data set. However, we have not yet explored the optimal solutions for camera visibility and coordinate partitioning. In some scenarios, we still require hyper-parameter tuning to provide better rendering quality, and our model relies on the accuracy of the initial sparse point cloud. Additionally, our research may be applied to the 3D mesh extraction in the large-scale scenes.These works are left for our future endeavors. 